%% file: main.tex
\documentclass[lettersize,journal]{IEEEtran}
\usepackage{amsmath,amsfonts}
\usepackage{algorithmic}
\usepackage{algorithm}
\usepackage{array}
\usepackage[caption=false,font=normalsize,labelfont=sf,textfont=sf]{subfig}
\usepackage{textcomp}
\usepackage{stfloats}
\usepackage{url}
\usepackage{verbatim}
\usepackage{graphicx}
\usepackage{cite}
\usepackage[breaklinks,colorlinks]{hyperref}

\usepackage{multirow}
\usepackage{makecell}
\usepackage{wrapfig}
\usepackage{xcolor}
\usepackage{colortbl}
\usepackage{tikz}
\usepackage{booktabs}

\usepackage[capitalize]{cleveref}
\crefname{paragraph}{paragraph}{paragraphs}
\Crefname{paragraph}{Paragraph}{Paragraphs}

\definecolor{falseNeg}{HTML}{3d9af5}
\definecolor{falsePos}{HTML}{fa5757}

\definecolor{prechange}{HTML}{F7B059}
\definecolor{postchange}{HTML}{5ADAFA}

\definecolor{irchange}{HTML}{D40808}
\definecolor{rchange}{HTML}{0816D4}

\definecolor{gold}{HTML}{FBF2D2}
\definecolor{silver}{HTML}{DDDDDD}
\definecolor{bronze}{HTML}{EED2B8}

\definecolor{goldD}{HTML}{D9AE13}
\definecolor{silverD}{HTML}{909090}
\definecolor{bronzeD}{HTML}{9A5F26}

\newcommand{\medal}[3]{\tikz[baseline=(char.base)]{\node[rounded corners=2pt,fill=#1,draw=#2,inner sep=1.5pt](char){#3};}}

\newcommand{\bm}[2]{
    \ifcase#1\or% case 1
      {\medal{gold}{goldD}{\textbf{#2}}}
    \or % case 2
      {\medal{silver}{silverD}{#2}}
    \or % case 3
      {\medal{bronze}{bronzeD}{#2}}
    \else % default case
      #2
    \fi\ignorespaces
}

\newcommand{\meanwithstd}[2]{\specialcell[c]{#1 \\[-3pt] {\scriptsize ($\pm$#2)}}}
\newcommand{\specialcell}[2][c]{%
  \begin{tabular}[#1]{@{}c@{}}#2\end{tabular}}

\definecolor{weights}{HTML}{0398fc}

\let\titleold\title
\renewcommand{\title}[1]{\titleold{#1}\newcommand{\thetitle}{#1}}
\def\maketitlesupplementary
   {
   \newpage
       \twocolumn[
        \centering
        \Large
        \textbf{\thetitle}\\
        \vspace{0.5em}Supplementary Material \\
        \vspace{1.0em}
       ] %< twocolumn
   }

\newcommand{\change}[1]{#1}

\begin{document}

\title{Make Some Noise: Unsupervised Remote Sensing Change Detection Using Latent Space Perturbations}


\author{Blaž Rolih, Matic Fučka, Filip Wolf, Luka Čehovin Zajc
\thanks{(Corresponding author: Blaž Rolih)}
\thanks{The authors are with the University of Ljubljana, Faculty of Computer and Information Science, 1000 Ljubljana, Slovenia
(e-mail: blaz.rolih@fri.uni-lj.si)}
}

\markboth{Preprint}%
{Rolih \MakeLowercase{\textit{et al.}}: Make Some Noise}


\maketitle

\begin{abstract}
Unsupervised remote sensing change detection (UCD) aims to localise changes between two images of the same region without relying on labelled training data. Most recent approaches either use a frozen foundation model in a training-free manner or train with synthetic changes generated in pixel space. Both strategies inherently rely on predefined assumptions about change types, typically introduced through handcrafted rules, external datasets, or auxiliary generative models. Due to these assumptions, such methods fail to generalise beyond a few change types, limiting their real-world usage, especially in rare or complex scenarios. 
To address this, we propose \textit{MaSoN} (\textbf{Ma}ke \textbf{So}me \textbf{N}oise), an end-to-end UCD framework that synthesises diverse changes directly in the latent feature space during training. It generates changes dynamically estimated from feature statistics of the target data, enabling diverse yet data-driven variation aligned with the target domain. Since synthesis happens in latent space, it also easily extends to new modalities, such as SAR and multispectral data. MaSoN generalises strongly across diverse change types and improves the average F1 score across five benchmarks by 14.1 percentage points. Project page: \url{https://blaz-r.github.io/mason_ucd/}
\end{abstract}


\section{Introduction}

\begin{figure}
    \centering
    \includegraphics[width=1\linewidth]{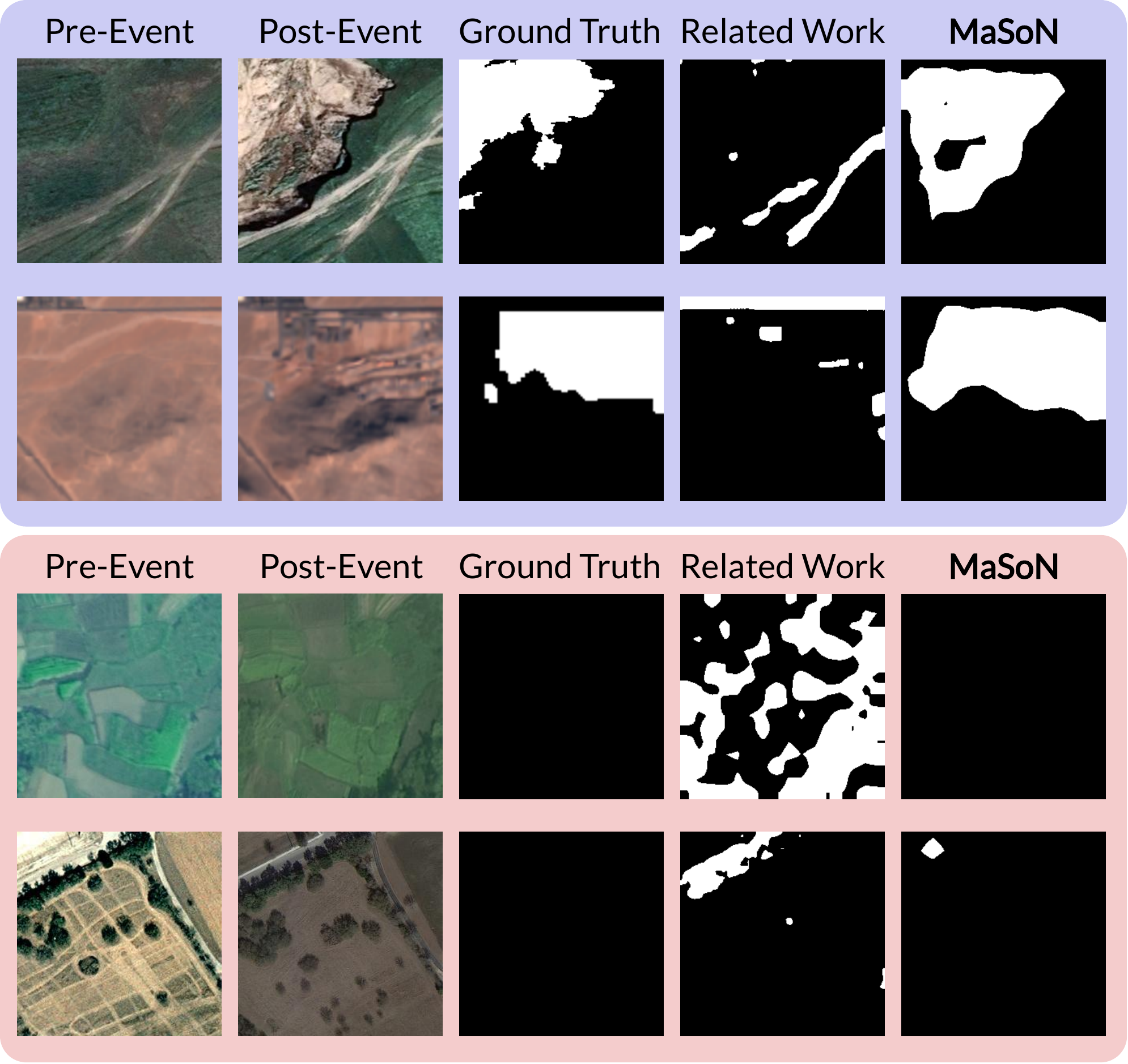}
    \caption{Related methods that rely on foundation models in a training-free manner or on changes generated in pixel-space often produce noisy or inaccurate masks, either missing \colorbox{rchange!30}{\textit{relevant changes}} or overreacting to \colorbox{irchange!30}{\textit{irrelevant changes}}. By generating changes in latent space, MaSoN learns from diverse and more data-aligned variations, producing better predictions with fewer false positives.
    This highlights MaSoN's improved ability to generalise across diverse and challenging change scenarios.}
    \label{fig:intro}
\end{figure}

Change detection (CD) is a fundamental task in remote sensing that involves localising changes in a sequence of images (typically two) of the same geographic region~\cite{daudt2018fcn, chen2021bit, ding2025labelEffsurvey, zheng2024anychange}. This task is essential to various applications, including disaster response, urban development monitoring, and land use mapping~\cite{daudt2018fcn, ding2025labelEffsurvey, zhu2022rsLandChange, hansch2024eo4climate}. Recent supervised methods have demonstrated impressive results~\cite{chen2021bit, daudt2018fcn, li2023a2net, chen2024changeMamba}, but they rely on pixel-level annotations and fail to generalise to unseen change types or new geographic contexts~\cite{ruuvzivcka2022ravaen, hertel2025rapid, bandara2023metricLearnCD, dietrich2025disaster}.

Dependence on labels presents a critical limitation for time-sensitive applications and situations where annotation is unfeasible. In these situations, unsupervised change detection (UCD) offers an effective alternative~\cite{ding2025labelEffsurvey, zheng2024anychange, ruuvzivcka2022ravaen}, as it removes the need for manual labelling and enables the use of large archives of unlabelled bi‑temporal imagery. By enabling fast analysis without human supervision, UCD methods support rapid response~\cite{hertel2025rapid, meneses2022rapidMap, valsamis2024wildfire, dietrich2025disaster} and efficient large‑area monitoring with reduced reliance on labour‑intensive annotation pipelines~\cite{ding2025labelEffsurvey}.

Recent UCD methods largely follow two strategies, each with limitations. \textit{Training free approaches}~\cite{zheng2024anychange, saha2020unsupervised, saha2019unsupervised, li2025dynearth} leverage frozen priors in a training-free manner (e.g., Segment Anything Model - SAM~\cite{kirillov2023sam}), but these are typically derived from natural/urban imagery and degrade under domain shift~\cite{hertel2025rapid} (e.g., landslides or croplands). State‑of‑the‑art \textit{learning-based methods} aim to overcome this by adopting a two-stage pipeline: they first \textit{generate} synthetic changes in pixel space and then \textit{train} a change detector on the generated samples~\cite{benidir2025hyscdg, chen2023_i3pe, noh2022cdrl, zheng2023changen}. 
However, the generated changes are shaped by assumptions introduced via auxiliary models~\cite{noh2022cdrl}, handcrafted rules~\cite{chen2023_i3pe, ding2025s2c}, or external datasets~\cite{zgeng2025changen2, zheng2021changeStar, benidir2025hyscdg}, which limits diversity and generalisation to novel change types.
Moreover,  methods from both families are in most cases restricted to RGB only input and often struggle with seasonal or radiometric variations (which we refer to as \textit{irrelevant changes}\footnote{Seasonal changes can be relevant in some applications, but we follow the majority of related work and benchmarks that consider them irrelevant.}).

We address these limitations with MaSoN (\textbf{Ma}ke \textbf{So}me \textbf{N}oise), a novel end-to-end UCD framework that trains on dynamically generated changes in the latent space. Instead of pixel-space augmentations, MaSoN injects Gaussian noise into the feature maps of a pretrained encoder. The noise scale is dynamically estimated using feature statistics from the target data, enabling on-the-fly generation of decoupled relevant and irrelevant changes. This design introduces diverse domain-data-aligned training signals without any labelled external data or a multi-stage setup. Because change synthesis occurs in feature space, MaSoN is modality‑agnostic and can be extended to new modalities such as multispectral and SAR with a simple encoder swap. As a result, MaSoN achieves superior performance across different change types and consistently outperforms prior methods across multiple datasets, as illustrated in \Cref{fig:intro}.

Our contributions are twofold: \textbf{(i)} \textit{We propose the first end-to-end latent space change generation and detection framework} that can be trained in an unsupervised manner with our on-the-fly change synthesis strategy. Our framework addresses the change-type generalisation problems in related approaches, thereby improving change detection performance across diverse change types.
\textbf{(ii)} \textit{We introduce a procedure for creating synthetic changes inside the latent space of an encoder. }
The synthetic changes are modelled as Gaussian noise, which is dynamically estimated using simple statistics derived from the latent features of the target data.
Through experiments and theoretical analysis, we show that real-world changes can be decoupled into irrelevant and relevant categories and approximated accordingly. This strategy enables variability that is hard to capture in pixel space, and it naturally supports non-RGB modalities \change{if a suitable encoder exists}.

We evaluate MaSoN on five benchmark datasets spanning diverse change types, including natural disasters, building changes, urban development, and cropland changes. MaSoN achieves superior performance on four out of five datasets, outperforming prior methods by an average of 14.1 percentage points, a relative increase of 38.6\%. We also demonstrate that MaSoN \textit{extends} and achieves strong results on multispectral and SAR modalities, which most related methods cannot easily do.

\section{Related work}
\label{s:rel}

\noindent\textbf{Remote sensing change detection} (CD) is a long-established task, with early approaches based on pixel-level comparisons and statistical techniques~\cite{singh1989reviewCD, le2013urbanSar, metzger2023UCForecast, peng2025deepDCSurvey}. Deep learning advancements have since enabled end-to-end CD using Siamese convolutional architectures~\cite{daudt2018fcn, li2023a2net, chen2020levirStanet} which were later extended to transformers~\cite{chen2021bit,bandara2022changeFormer,yu2024maskcd,zhang2022swinsunet}, state-space models like Mamba~\cite{chen2024changeMamba, gu2023mamba, liu2024vmamba} and diffusion-based approaches~\cite{bandara2025ddpmcd, benidir2025hyscdg, zgeng2025changen2}. However, most recent models remain supervised and heavily rely on scarce annotated datasets, which are difficult to collect, especially for rare events such as natural disasters~\cite{ding2025labelEffsurvey, hertel2025rapid}. Furthermore, such models often fail to generalise to unseen scenarios~\cite{ruuvzivcka2022ravaen, hertel2025rapid}, requiring costly supervised retraining.

\noindent\textbf{Unsupervised change detection} addresses these limitations by operating solely on unannotated image pairs, making it especially valuable for rapid disaster response and other time-critical applications where annotated data is unavailable~\cite{hertel2025rapid, ruuvzivcka2022ravaen, zheng2024anychange}.
Methods like image-differencing~\cite{singh1989reviewCD}, Markov Random Fields~\cite{bruzzone2000automatic}, and change vector analysis (CVA)~\cite{bovolo2006theoretical, bruzzone2000automatic} rely on probability theory and statistics. While historically important, they are limited in expressiveness and are outperformed by modern deep-learning counterparts~\cite{saha2020unsupervised, saha2019unsupervised, du2019sfa, zheng2024anychange}.

\change{Another branch of unsupervised CD additionally considers image pairs acquired by different sensors (heterogeneous CD). IRG-McS~\cite{sun2021irgmcs} represents the images as graphs and detects changes through cross-image graph mappings, whereas RIEM~\cite{sun2025riem} infers changes through a rules-induced energy model. These methods explicitly mitigate modality differences through designed structural or relational constraints.}

Several recent UCD approaches leverage large vision foundation models (VFMs). AnyChange~\cite{zheng2024anychange}, SCM~\cite{tan2024scm}, and DynamicEarth~\cite{li2025dynearth} use Segment Anything Model (SAM)~\cite {kirillov2023sam} for segmentation, and CDRL-SA~\cite{noh2024cdrlSA} uses it for refinement. Although effective in some scenarios, these models rely on SAM with frozen prior knowledge from the natural image domain, making them less suitable for remote sensing tasks. In contrast, our method is learning-based, allowing representations to adapt to the domain, thus improving generalisation and flexibility.

Other learning-based methods create synthetic changes in pixel-space~\cite{zgeng2025changen2, song2024syntheworld, chen2023_i3pe, noh2022cdrl}. ChangeStar~\cite{zheng2021changeStar} constructs synthetic data from unrelated segmentation datasets. Other methods, such as CDRL~\cite{noh2022cdrl}, FCD-GAN~\cite{wu2023fcdgan}, SyntheWorld~\cite{song2024syntheworld}, Changen~\cite{zheng2023changen, zgeng2025changen2}, and HySCDG~\cite{benidir2025hyscdg} use generative models for style transfer or full image synthesis. These models require auxiliary external datasets to generate synthetic changes, which limits diversity and domain generalisation. I3PE~\cite{chen2023_i3pe} and S2C~\cite{ding2025s2c} remove such dependencies but still operate in pixel space using handcrafted transformations, leading to constrained variability. In contrast, MaSoN generates synthetic changes directly in the latent feature space during training. These changes are dynamically estimated using feature statistics of the target data, enabling diverse yet domain-consistent changes and stronger generalisation.

\noindent\textbf{Learning with Noise} is an established strategy in deep learning for improving robustness and generalisation. DN-DETR~\cite{li2022dndetr} uses query denoising to accelerate training and stabilise bipartite matching for \textit{object detection}. In \textit{anomaly detection}, noise was applied to simulate anomalies~\cite{liu2023simplenet, rolih2024supersimplenet, rolih2025ssn2}. GeoCLIP~\cite{vivanco2023geoclip} leverages noise to model GPS inaccuracies and for contrastive learning in \textit{geolocalisation}.
It has also been applied to mitigate label uncertainty in hyperspectral change detection~\cite{li2019unsupNoiseModeling}. Building on these foundations, MaSoN introduces a novel strategy to generate changes with Gaussian noise directly in the feature space. Unlike previous methods that rely on fixed~\cite{liu2023simplenet, pang2019devNet} and non-decoupled~\cite{cai2022perturbation} Gaussian noise, MaSoN decouples it into irrelevant and relevant noise and dynamically adjusts it during training to address the inherent variability in remote sensing change detection.

\begin{figure*}[!t]
    \centering
    \includegraphics[width=1\linewidth]{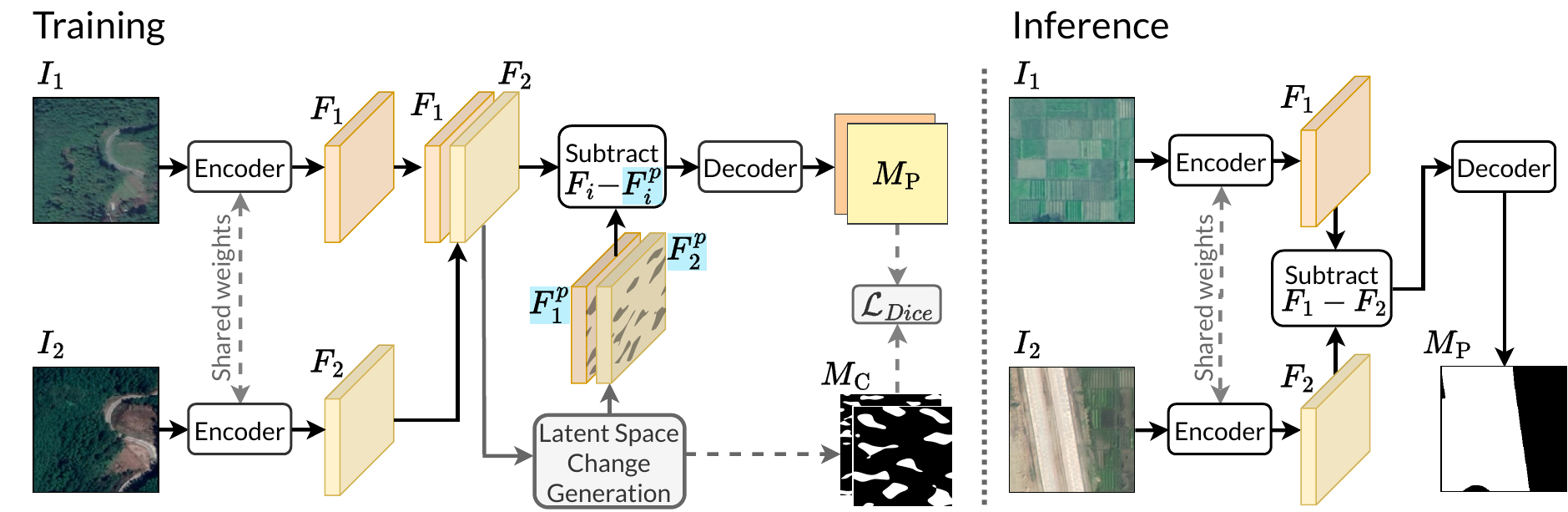}
    \caption{The architecture of the proposed unsupervised change detection framework, MaSoN.}
    \label{fig:arch}
\end{figure*}

\section{Let's Make Some Noise}

The proposed method (MaSoN) consists of a \textit{Shared Weight Encoder}, a \textit{Latent Space Change Generation Strategy}, and a \textit{Mask Decoder}. 
The features are first extracted using the \textit{Encoder} ($\Phi$).
Then the \textit{Latent Space Change Generation Strategy} is used (only during training) to generate synthetic changes on the feature level. Features from different time steps with generated synthetic changes are then fused (in our case via elementwise subtraction), and fed into the \textit{Mask Decoder} ($\mathbb{D}$), which outputs the predicted change mask. The architecture of our method is depicted in \Cref{fig:arch} and explained in the following subsections.

\subsection{Encoder}
\label{sub:feat_extr}

Given a bi-temporal input image pair ($I_1, I_2$), we extract features ($F_1$, $F_2$) using a shared-weight pretrained encoder $\Phi$. 
For each image $I_i$, the encoder produces a hierarchical feature set:
\begin{equation}
    \Phi(I_i) = F_i = \{ f_i^{(l)}\; |\; l \in L \},
\end{equation}
where $l$ denotes the layer level 
and $f_i^{(l)}$ the feature map at layer $l$ from image $I_i$.

\subsection{Why Gaussian noise?}
\label{sub:feat_analysis}

To better understand and motivate our synthetic change generation procedure, we analyse the behaviour of features extracted by the pretrained encoder. Our goal is to determine whether changes can be meaningfully characterised and thus approximated directly in the latent space.

\begin{figure*}[!h]
    \centering
    \includegraphics[width=1\linewidth]{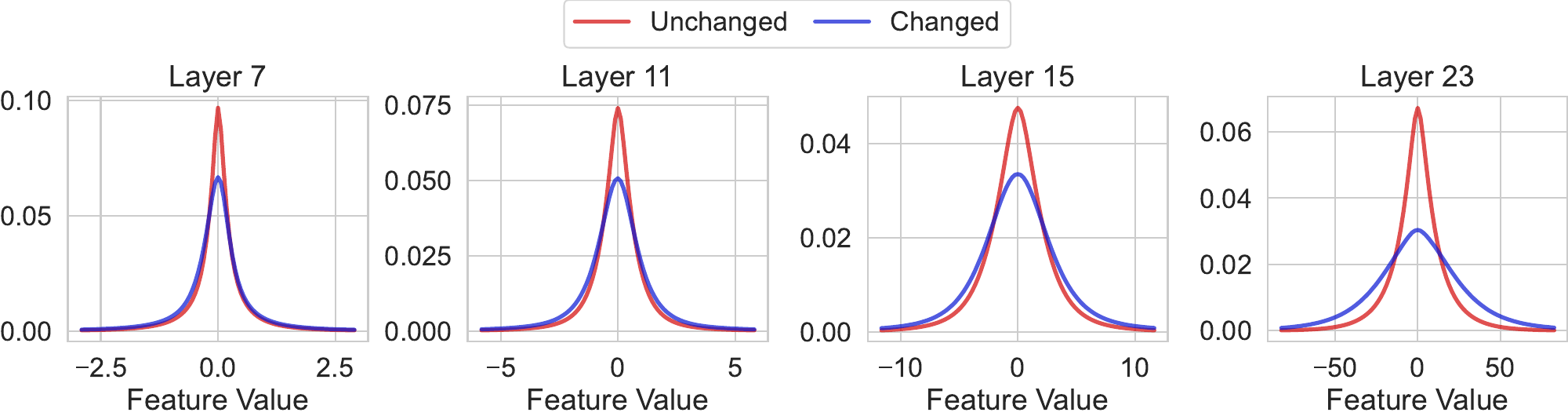}
    \caption{Histogram plot of feature differences $f_1^{(l)} - f_2^{(l)}$, averaged across all five datasets and all channels per layer. \colorbox{irchange!30}{Unchanged} regions are narrowly concentrated near zero, while \colorbox{rchange!30}{changed} regions exhibit broader variation, especially in deeper layers. Both distributions can be approximated by a zero-centred Gaussian, but with different variance parameters. This directly motivates our latent-space change generation strategy.}
    \label{fig:feats}
\end{figure*}

We perform the analysis on five different benchmark datasets (described later in~\Cref{par:metr_data}).
For each image pair, we extract features and compute element-wise difference at each layer: $f_{diff}^{(l)} = f_1^{(l)} - f_2^{(l)}$. Using the ground truth annotations, we group these differences into \colorbox{rchange!30}{changed} and \colorbox{irchange!30}{unchanged} regions. Figure~\ref{fig:feats} shows histograms of these difference values averaged across all datasets for each layer in the feature hierarchy. 

\noindent \textbf{Main Takeaways.} According to our analysis, changed and unchanged regions exhibit distinct patterns in feature space. Unchanged areas produce small differences, resulting in sharp peaks near zero with a narrow distribution. Changed regions show a broader, heavier-tailed distribution, with fewer near-zero differences, especially in deeper layers that capture high-level semantic concepts. Although both distributions overlap and are \textit{centred at zero}, their \textit{variances differ}.
From a theoretical perspective, the \emph{maximum-entropy} principle states that, among all distributions consistent with the known constraints, the one with the largest entropy is the best representation of our current knowledge~\cite{jaynes1957maxEnt}. Given the observed zero mean and per-channel variances, the \textit{Gaussian distribution satisfies these constraints} and \textit{maximises differential entropy}~\cite{cover2005infoTheo}.

The theoretical grounding and empirical similarity to the Gaussian distribution suggest that both unchanged and changed regions can be synthetically generated (or rather \change{locally} approximated) during training using properly calibrated Gaussian noise. We later demonstrate that calibration can be achieved using batch statistics \change{in order to adapt to diverse remote sensing scenes.}
An extended theoretical justification is provided in Supplementary E, while full empirical implementation details, per-dataset results, and discussion are given in Supplementary C.

\subsection{Latent Space Change Generation Strategy}
\label{sub:changen}

\begin{figure}[!h]
    \centering
    \includegraphics[width=1\linewidth]{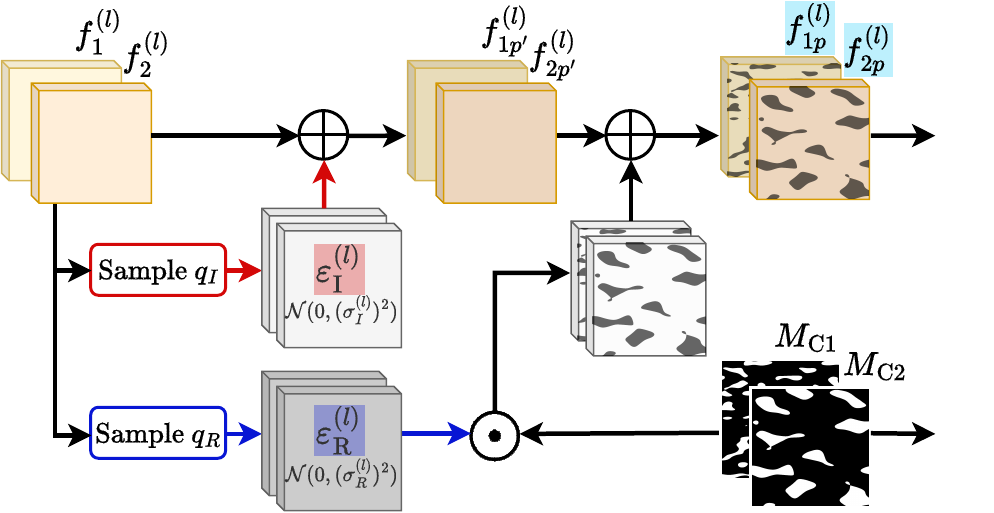}
    \caption{Illustration of the change generation process.}
    \label{fig:changen}
\end{figure}

As illustrated in \cref{fig:changen}, given feature map sets of an image pair, $F_1$ and $F_2$, our goal is to synthesise training pairs with controllable changes.
Since image pairs may already contain (unknown) changes, directly using each pair in a bi-temporal fusion and generating additional synthetic changes would cause the model to ignore unknown changes. Instead, we treat each image independently and generate synthetic changes by perturbing its own features.

We divide changes between images into two categories:
(i) \colorbox{irchange!30}{\textit{irrelevant changes}}, such as illumination shifts, minor vegetation growth, or seasonal changes, and
(ii) \colorbox{rchange!30}{\textit{relevant changes}}, which correspond to large changes like building construction or landslide damage.
Based on our analysis in Subsection~\ref{sub:feat_analysis}, we simulate these two types of changes using zero-centred Gaussian noise with a separate variance parameter, \change{which is dynamically estimated to adapt to diverse scenes.} 

\noindent\colorbox{irchange!30}{\textbf{Irrelevant Change Noise.}} Empirically, the distribution of feature differences between unchanged regions follows a Gaussian-like shape (see \cref{sub:feat_analysis}). Motivated by this and the maximum entropy principle (see \Cref{sub:feat_analysis}), we model irrelevant changes using Gaussian noise. Specifically, we sample a noise \colorbox{irchange!30}{$\varepsilon_{\text{I}}$}:
\begin{equation}
\colorbox{irchange!30}{$\varepsilon_{\text{I}}^{(l)}$} \sim \mathcal{N}(0, (\sigma^{(l)}_I)^2),
\label{eq:irrel}
\end{equation}
where the $\sigma^{(l)}_I$ is set to the $q_I$-th quantile of the absolute feature difference of specific feature layer $l$:
\begin{equation}
\sigma^{(l)}_I = \text{Quantile}(|f_1^{(l)} - f_2^{(l)}|, q_I).
\end{equation}
$q_I$ is a learnable parameter, which is initialised with a lower value (in our case, 0.85) to approximate the narrow distribution observed in our analysis.
Quantile sampling is performed dynamically for each layer and feature channel within every training batch, allowing the noise magnitude to adapt to local feature statistics. The \textit{key idea} is sampling from feature difference, which enables the extraction of irrelevant variation directly from the feature differences. This enables estimation of intra-image irrelevant variation without relying on fixed values.

Without injecting \colorbox{irchange!30}{$\varepsilon_{\text{I}}$}, the model would associate unchanged regions with exactly zero feature difference, making it overly sensitive to minor variations at test time.

\noindent\colorbox{rchange!30}{\textbf{Relevant Change Noise.}} Similarly, motivated by our feature analysis in \Cref{sub:feat_analysis}, the distribution of feature differences in changed regions approximately follows a Gaussian pattern. To approximate this, noise \colorbox{rchange!30}{$\varepsilon_{\text{R}}$} is sampled as:
\begin{equation}
\colorbox{rchange!30}{$\varepsilon_{\text{R}}^{(l)}$} \sim \mathcal{N}(0, (\sigma^{(l)}_R)^2),
\end{equation}
where $\sigma^{(l)}_R$ is the $q_R$-th quantile computed from the concatenation of both feature sets along the batch dimension:
\begin{equation}
\sigma^{(l)}_R = \text{Quantile}(\text{Concat}([f_1^{(l)}, f_2^{(l)}], \text{dim} = \text{batch}), q_R).
\end{equation}
We sample quantiles from both feature sets (concatenation) so the sampled value stays within the realistic bounds. $q_R$ is a learnable parameter initialised to a value close to one (in our case, 0.98), so it captures the broader variance observed for relevant (changed) regions. \change{Higher quantile initialisation paired with a low learning rate also encourages that the irrelevant and relevant noises do not converge to the same value.}
As with irrelevant noise, the quantiles are sampled independently per channel each iteration.

\change{Since the model operates in a binary setting, this variation captures all major changes, without explicitly differentiating the change class (i.e. building, landslide). We demonstrate extension to type-specific filtering at the end of SubSec~\ref{sub:res}.}

The noise is applied selectively using a binary mask $M_{\text{C}}$, generated from thresholded Perlin noise~\cite{perlin1985image} (see \Cref{fig:changen}). The mask ensures that the changed region is random but spatially consistent, while conveniently serving as the ground truth during training.

\begin{figure}[!h]
    \centering
    \includegraphics[width=1\linewidth]{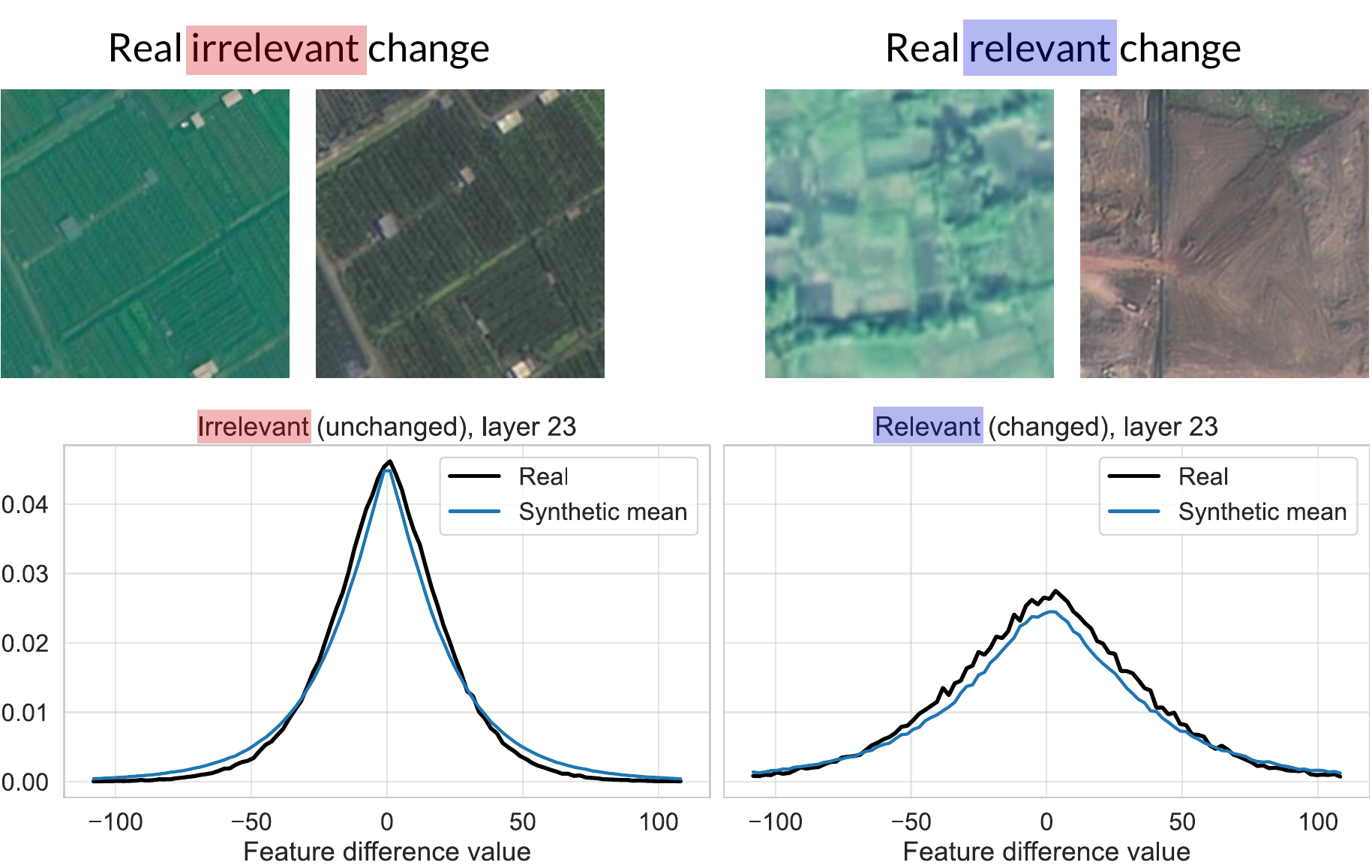}
    \caption{\change{Illustration of true irrelevant (unchanged) and relevant (changed) changes from CLCD~\cite{li2022clcdMSCANET} (top). Histograms comparing real and synthetic feature-difference distributions (bottom). Our approach approximates the true distribution of feature differences between changed and unchanged regions.}}
    \label{fig:noise_fig}
\end{figure}

\change{Figure~\ref{fig:noise_fig} visually illustrates an example of real relevant (changed) and irrelevant (unchanged) variation from CLCD~\cite{li2022clcdMSCANET}. The bottom part of the figure compares feature differences between the real example and our synthetic approach, obtained by randomly sampling Gaussian noise using the above approach 10 times. The simulated perturbations sufficiently approximate the true distributions of irrelevant and relevant variation for the purpose of training.}

\noindent\textbf{Final Synthetic Pairs.} We generate the final synthetic training pairs (also depicted in \cref{fig:changen}) for each layer as:
\begin{equation}
\begin{aligned}
    &(f_1^{(l)}, f_1^{(l)} + \colorbox{irchange!30}{$\varepsilon_{\text{I}}^{(l)}$} + M_{\text{C}1} \odot \colorbox{rchange!30}{$\varepsilon_{\text{R}}^{(l)}$})\\
    &(f_2^{(l)}, f_2^{(l)} + \colorbox{irchange!30}{$\varepsilon_{\text{I}}^{(l)}$} + M_{\text{C}2} \odot \colorbox{rchange!30}{$\varepsilon_{\text{R}}^{(l)}$})\\
    &=(f_1^{(l)}, \colorbox{postchange!30}{$f_{1p}^{(l)}$})(f_2^{(l)}, \colorbox{postchange!30}{$f_{2p}^{(l)}$}),
\end{aligned}
\label{eq:changen}
\end{equation}
where $\odot$ denotes element-wise multiplication. The synthetic training pairs are then assembled back into a set across all layers, resulting in two training pairs: 
\begin{equation}
\begin{aligned}
&\Bigl(\{f_1^{(l)}  \; |\; l \in L\}, \colorbox{postchange!30}{$\{f_{1p}^{(l)}  \; |\; l \in L\}$}\Bigr)\\
&\Bigl(\{f_2^{(l)}  \; |\; l \in L\}, \colorbox{postchange!30}{$\{f_{2p}^{(l)}  \; |\; l \in L\}$}\Bigr)\\ 
&=(F_1, \colorbox{postchange!30}{$F_1^{p}$})\
(F_2, \colorbox{postchange!30}{$F_2^{p}$}),
\end{aligned}
\label{eq:changen_final}
\end{equation}
where \colorbox{postchange!30}{$F_i^p$} denotes the synthetic perturbed feature set. This formulation enables the model to learn to detect relevant semantic changes while remaining robust to irrelevant variations, all without requiring any labelled change annotations. 

Unlike prior works from other fields (e.g., anomaly detection) where noise is used as a static~\cite{liu2023simplenet, pang2019devNet, rolih2024supersimplenet}, non-decoupled~\cite{cai2022perturbation} or robustness‑oriented~\cite{li2022dndetr, vivanco2023geoclip} perturbation, MaSoN \emph{dynamically} estimates two \emph{decoupled} noise components (irrelevant and relevant) from feature statistics, a design motivated by the higher intra‑ and inter‑image variability of remote sensing data.

\subsection{Mask Decoder}

To produce the final change mask prediction, feature pairs are first subtracted elementwise:
\begin{equation}
\begin{aligned}
        &F^{diff}_i = F_i - \colorbox{postchange!30}{$F_i^p$} \quad \text{during \textit{training},} \quad \text{and} \quad \\
        &F^{diff} = F_1 - F_2 \quad \text{during \textit{inference}.}
\end{aligned}
\end{equation}

The resulting set of feature differences is passed to the UPerNet decoder~\cite{xiao2018upernet}, which processes a hierarchy of feature differences to produce the final change map $M_{\text{P}i}$, which is of the same spatial dimensions as the input image:

\begin{equation}
    M_{\text{P}i} = \mathbb{D}(F^{diff}_i) \enspace .
    \label{eq:mp}
\end{equation}

Since we make two training pairs from a single input pair during training (\cref{eq:changen,eq:changen_final}), the decoder also predicts two maps $M_{\text{P}1}$ and $M_{\text{P}2}$. During inference, the decoder produces a single map per feature pair ($M_\text{P}$, \cref{fig:arch} right).

\subsection{Training and Inference}

The model is trained using Dice loss~\cite{Milletari2016diceVnet} where the prediction is the change map $M_{\text{P}i}$ from \cref{eq:mp} and the target is set to the binary mask $M_{\text{C}i}$ from \cref{eq:changen}:
\begin{equation}
    \mathcal{L} = \mathcal{L}_{Dice}(M_{\text{P}1}, M_{\text{C}1}) + \mathcal{L}_{Dice}(M_{\text{P}2}, M_{\text{C}2}) \enspace .
\end{equation}

During inference, the change generation mechanism is removed, and the model directly predicts the change map from a pair of images (\cref{fig:arch} right). To obtain the binary mask, a sigmoid and a threshold of 0.5 are applied.

\section{Experiments}

\paragraph{Implementation Details.}
\label{par:impldet}

As the encoder, we use the ViT-L~\cite{dosovitskiy2021vit} version of DINOv3~\cite{simeoni2025dinov3}, which we do not finetune. Following related work~\cite{wang2024mtp}, we take features from layers 7, 11, 15, and 23. Input images are cropped to $256 \times 256$ pixels.
During training, we apply the flip and rotate augmentations. The model is trained using the AdamW~\cite{Loshchilov2017adamw} optimiser, with an initial learning rate of $10^{-5}$ for the decoder and $10^{-7}$ for the learnable quantile parameters (2 values, one for $q_I$ and one for $q_R$). A cosine scheduler without restarts is used. Each training run lasts 1000 iterations with a batch size of 16 and completes in approximately 7 minutes on a single NVIDIA A100 GPU. Additional details can be found in the Supplementary N.

\paragraph{Evaluation Metrics and Datasets.}
\label{par:metr_data}

Following standard practice~\cite{bandara2022changeFormer, chen2021bit, daudt2018fcn, zheng2023changen, zheng2024anychange, rolih2025btc}, we evaluate change detection performance using \textbf{binary} precision, recall, and F1 score considering only \textbf{change class}. Experiments are repeated across five random seeds, and we report the mean and standard deviation. For each run, the model checkpoint from the final epoch is used. For training-free baselines, we use author-provided weights.

To ensure robustness, we evaluate on five change detection datasets covering \textit{worldwide locations} and spanning \textit{diverse change types}, from \textit{building} and \textit{urban} changes to \textit{cropland} and \textit{natural disaster} changes. The datasets cover different ground sample distances, sensors, and global geographic regions. \textit{SYSU}~\cite{shi2022sysuDSAMnet} includes multiple change types such as buildings, groundwork, vegetation, and sea changes. \textit{LEVIR}~\cite{chen2020levirStanet} focuses on building changes, while \textit{CLCD}~\cite{li2022clcdMSCANET} captures cropland changes. We also include datasets that are particularly challenging for unsupervised methods: \textit{GVLM}~\cite{zhang2023gvlm}, containing natural disaster changes resulting from landslides, and \textit{OSCD}~\cite{daudt2018urban}, a low-resolution Sentinel-2 dataset covering worldwide urban changes. The models are trained on the dedicated training sets (using no labels) and evaluated on the corresponding test sets. Additional dataset details are provided in Supplementary B.

\subsection{Experimental Results}
\label{sub:res}

\noindent\textbf{Unsupervised Change Detection.} We compare MaSoN against a range of unsupervised change detection methods. We include image pixel differencing~\cite{singh1989reviewCD, hussain2013cdPixToObj}, DCVA~\cite{saha2019unsupervised}, and DINOv3~\cite{simeoni2025dinov3} based CVA~\cite{zheng2024anychange}. We include state-of-the-art methods HySCDG~\cite{benidir2025hyscdg}, I3PE~\cite{chen2023_i3pe}, Changen2~\cite{zgeng2025changen2}, AnyChange~\cite{zheng2024anychange}, SCM~\cite{tan2024scm}, DynamicEarth~\cite{li2025dynearth} and S2C~\cite{ding2025s2c}. Their implementation details are provided in the Supplementary P. Results are summarised in Table~\ref{tab:sota}.

\begin{table*}[!ht]
    \caption{Unsupervised change detection results across five diverse datasets and their average. We report Precision (Pr.), Recall (Re.), and F1 score. Results of additional metrics for individual datasets are provided in the Supplementary E. FPS was benchmarked on an NVIDIA A100 (details and more computation metrics are in Supplementary H). \textcolor{goldD}{First} and \textcolor{silverD}{second} place results are marked.}
    \label{tab:sota}
\resizebox{\linewidth}{!}{
\setlength{\tabcolsep}{3pt}
    \centering
    \begin{tabular}{lcc|ccccc|ccc}
    \toprule
    	\multirow{2}{*}{~} & FPS & Param.& SYSU& LEVIR& GVLM& CLCD& OSCD& \multicolumn{3}{c}{\textit{Average}} \\
~ & [img/s] & [M]& F1& F1& F1& F1& F1& Pr. & Re. & F1\\
 \hline
Pixel Difference~\cite{singh1989reviewCD} & 456.4{\scriptsize$\pm54.5$} & 0.0 & 42.9 & 8.9 & 23.4 & 16.1 & 18.5 & 16.2 & 49.3 & 22.0\\
DCVA~\cite{saha2019unsupervised}\tiny{TGRS19} & 0.4{\scriptsize$\pm0.0$} & 0.5 & 2.5 & 0.2 & 3.2 & 3.1 & \bm2{ 38.9} & 30.0 & 9.3 & 9.6\\
DINOv3-CVA~\cite{zheng2024anychange} & 24.4{\scriptsize$\pm0.3$} & 303.1 & 56.5 & 19.1 & 19.3 & 27.1 & 21.3 & 19.0 & \bm1{ 80.4} & 28.7\\
AnyChange~\cite{zheng2024anychange}\tiny{NeurIPS24} & 0.4{\scriptsize$\pm0.0$} & 641.1 & 46.0 & 25.4 & 20.8 & 26.5 & 26.1 & 26.0 & 43.0 & 29.0\\
Changen2-S9~\cite{zgeng2025changen2}\tiny{TPAMI25} & 33.3{\scriptsize$\pm0.1$} & 99.9 & 44.8 & 19.9 & 19.1 & 15.6 & 1.5 & 17.3 & 39.4 & 20.2\\
I3PE~\cite{chen2023_i3pe}\tiny{ISPRS23} & 65.9{\scriptsize$\pm0.2$} & 24.9 & 33.7 & 16.8 & 20.3 & 11.5 & 3.6 & 15.6 & 24.8 & 17.2\\
HySCDG~\cite{benidir2025hyscdg}\tiny{CVPR25} & 41.0{\scriptsize$\pm0.1$} & 65.1 & 52.7 & 15.4 & 17.3 & 24.9 & 17.2 & 29.2 & 65.2 & 25.5\\
SCM~\cite{tan2024scm}\tiny{IGARSS24} & 1.2{\scriptsize$\pm0.0$} & 223.5 & 25.7 & 27.3 & 28.0 & 18.9 & 16.6 & 21.6 & 30.1 & 23.3\\
DynEarth~\cite{li2025dynearth}\tiny{AAAI26} & 0.3{\scriptsize$\pm0.0$} & 877.6 & 56.2 & \bm2{ 46.2} & 16.6 & 29.5 & 26.6 & 30.8 & 45.9 & 35.0\\
DynE. (DINOv3)~\cite{li2025dynearth}\tiny{AAAI26} & 0.3{\scriptsize$\pm0.0$} & 877.6 & \bm2{ 60.3} & \bm1{ 49.5} & 19.3 & 33.7 & 18.2 & 34.9 & 42.0 & 36.2\\
S2C (DINOv3)~\cite{ding2025s2c}\tiny{AAAI26} & 3.8{\scriptsize$\pm0.0$} & 303.6 & 37.6 & 32.0 & \bm2{ 38.1} & \bm2{ 52.2} & 5.1 & \bm2{ 41.4} & 42.1 & \bm2{ 36.5}\\
\textbf{MaSoN} & 39.7{\scriptsize$\pm0.1$} & 337.5 & \meanwithstd{\bm1{60.6}}{2.2} & \meanwithstd{38.9}{1.2} & \meanwithstd{\bm1{55.4}}{2.9} & \meanwithstd{\bm1{54.3}}{1.5} & \meanwithstd{\bm1{43.5}}{1.0} & \meanwithstd{\bm1{44.4}}{1.9} & \meanwithstd{\bm2{67.2}}{3.8} & \meanwithstd{\bm1{50.6}}{0.4}\\

    \bottomrule
    \end{tabular}
    }
\end{table*}

MaSoN achieves the highest average F1 score of 50.6, surpassing the previous best S2C~\cite{ding2025s2c} by 14.1 percentage points (p.\ p.) on average (a 45.41\% relative gain). S2C, which trains by generating changes in pixel spaces using augmentations, performs well on CLCD, it degrades substantially in other settings. We attribute MaSoN’s better performance to \textit{data-grounded latent-space change synthesis}, which produces diverse, realistic semantic variations directly in a more robust latent representation space.

\noindent\textbf{Comparison against training-free methods.} AnyChange~\cite{zheng2024anychange}, SCM~\cite{tan2024scm}, and DynamicEarth~\cite{li2025dynearth}, all built on SAM~\cite{kirillov2023sam}, underperform MaSoN on CLCD, GVLM, and OSCD. DynamicEarth exceeds MaSoN only on LEVIR (building change detection) as it produces more precise building borders (see qualitative results discussion), consistent with SAM’s strength in object-centric scenes. Overall, this suggests that training-free SAM does not always transfer well to remote sensing regimes that are less object-centric and/or lower-resolution, where foreground--background cues are weaker than in natural images.

\noindent\textbf{Comparison against pixel-space change generation.}
Pixel-space synthesis is consistently less effective than latent-space generation. S2C~\cite{ding2025s2c} and I3PE~\cite{chen2023_i3pe} results suggest that hand-crafted pixel transformations cannot capture the bre\-adth of possible changes. Even deep generative pixel-space methods (Chan\-gen2~\cite{zgeng2025changen2}, HySCDG~\cite{benidir2025hyscdg}) remain far behind MaSoN, supporting the view that pixel-level synthetic data generalises poorly across diverse change types and acquisition conditions. 

\noindent\textbf{Comparison against traditional methods.}
Compared to change vector analysis (CVA)~\cite{bruzzone2000automatic} computed on features from the same DINOv3~\cite{simeoni2025dinov3} encoder, MaSoN doubles precision and improves average F1 by 21.9 p.p. This confirms that strong representations help, but MaSoN’s gains are not due to the backbone alone. The learnable pipeline enabled by latent-space change synthesis is essential for turning features into accurate unsupervised change detection.

\begin{figure}[!h]
    \centering
    \includegraphics[width=1\linewidth]{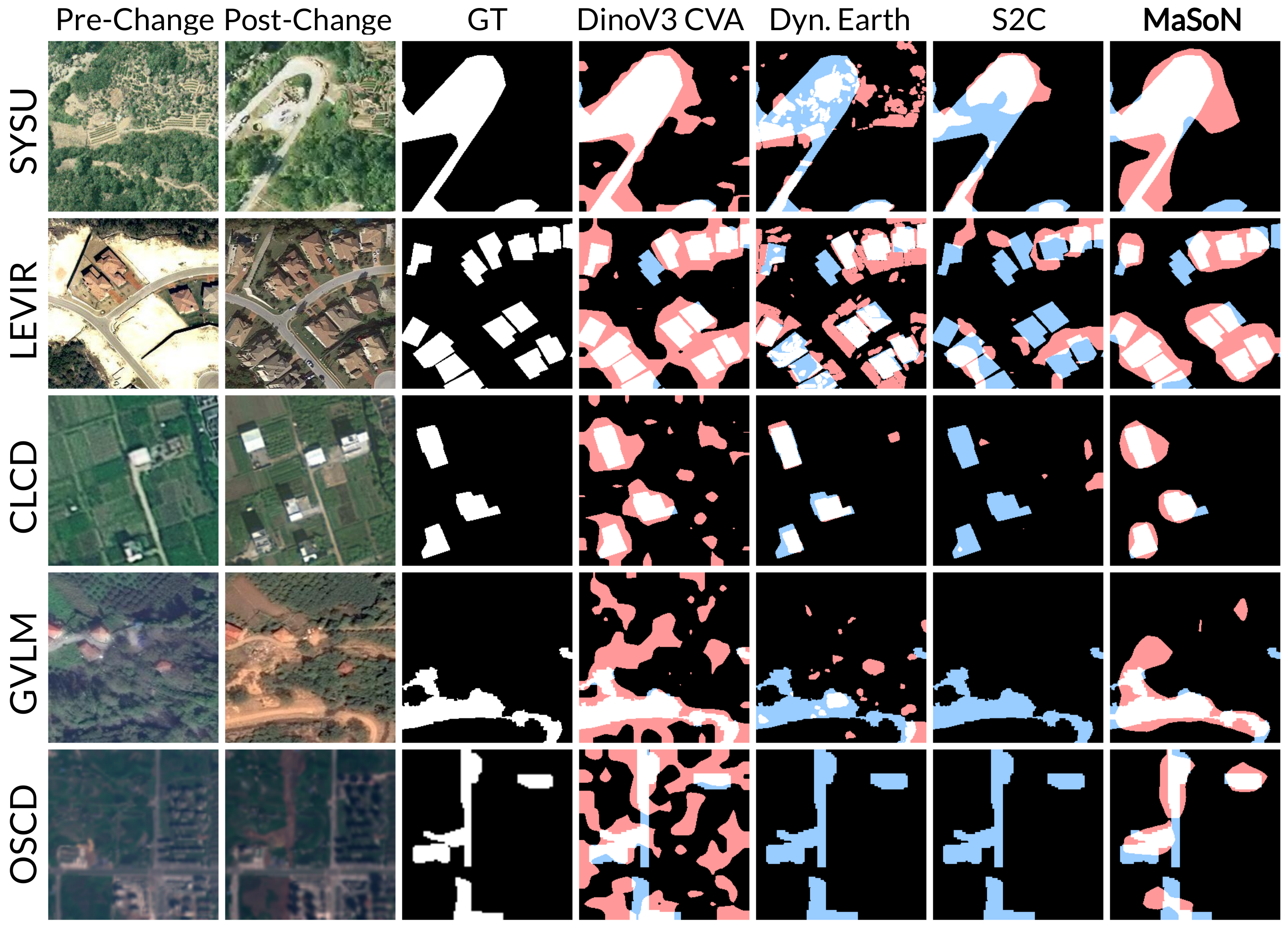}
    \caption{Qualitative comparison of the predictions. The pair of considered images is shown in the first and second columns, followed by the ground truth mask and predictions for each method. \textcolor{falsePos}{False positives are marked in red} and \textcolor{falseNeg}{false negatives in blue}. Additional qualitative results, including discussion and failure cases, are presented in Supplementary G.}
    \label{fig:qual}
\end{figure}

\noindent\textbf{Qualitative examples.} Figure~\ref{fig:qual} illustrates MaSoN's strong generalisation a\-cross diverse change types, particularly on GVLM (landslides) and OSCD (low-resolu\-tion), where previous SOTA methods DynamicEarth and S2C fail. On LEVIR, MaSoN occasionally overextends building regions, an issue shared by most methods except DynamicEarth, which uses SAM and is more biased toward object-centric detection. While DINOv3-based CVA captures most changes, it does so at the cost of excessive false positives, an issue that MaSoN substantially reduces.

\noindent\textbf{Extension to Specific Changes Filtering.}
Certain applications may require isolating or filtering specific types of changes. To address this, MaSoN can be extended with a text-guided filtering procedure similar to one from DynamicEarth~\cite{li2025dynearth} (see Supplementary R). To evaluate its performance in this scenario, we run the experiment on the LEVIR~\cite{chen2020levirStanet} dataset, strictly focusing on buildings. The performance \textit{improves from} 38.9\% F1 \textit{to} 50.3\% F1, surpassing the original DynamicEarth, which achieves 49.5\%. As seen in \Cref{tab:sota}, MaSoN already outperforms DynamicEarth on all other datasets without filtering. We experimented with filtering across other datasets; however, this did not yield consistent improvements. We hypothesise that this comes from the complex nature of many change types (e.g., landslides or cropland), which are difficult to describe unambiguously by text. Consequently, we evaluated filtering on LEVIR, as building changes are textually well-defined. These findings highlight both MaSoN's flexibility and the current limitation of text-based filtering for domain-specific changes.

\noindent\textbf{Extension to a new modality.}
The RGB modality can be unreliable in certain UCD scenarios, for example, for cloud‑covered flood imagery. MaSoN can be directly applied to SAR data by replacing the RGB encoder with a SAR‑capable alternative, such as Copernicus‑FM~\cite{wang2025copfm}. On the Sentinel‑1 OMBRIA flood‑change dataset~\cite{drakonakis2022ombrianet} (\Cref{tab:ombria}), MaSoN outperforms both pixel differencing and CVA when using Copernicus‑FM features. Other methods in \Cref{tab:sota} cannot be easily extended to SAR, as they rely on RGB‑only SAM or restrict their change‑genera\-tion mechanisms to the RGB domain. 

We also extend MaSoN to the multispectral Sentinel-2 dataset using the DEO~\cite{wolf2026deo} backbone and evaluate it on OSCD-multispectral. It achieves an F1 score of 45.1, which is better than RGB-MaSoN and all related methods that process only the RGB part of OSCD (see \Cref{tab:sota}).

\change{We further evaluate MaSoN on heterogeneous optical--SAR change detection. We use co-registered Landsat-8 optical and Sentinel-1A SAR observations of flooding in California from~\cite{lupino2019hetData} and process both with Copernicus-FM~\cite{wang2025copfm} to obtain features. MaSoN obtains 22.56 F1, compared with 12.84 F1 for feature CVA using the same Copernicus-FM encoder. Thus, the learned latent-perturbation framework provides a clear improvement over direct feature comparison, but the absolute heterogeneous performance remains limited compared to rule-based methods like RIEM~\cite{sun2025riem} which achieves 49.9 F1.}

\begin{table}[!t]
    \setlength{\tabcolsep}{3pt}
    \centering
                \caption{Results on the OMBRIA Sentinel-1 SAR flood change-detection dataset.}
            \label{tab:ombria}
        \begin{tabular}{lccc}
            \toprule
        Method & Precision & Recall & F1 \\
                \midrule
                Pixel Diff.  & \change{35.96} & \change{43.33} & 39.98 \\
                Cop.-FM CVA  & \change{46.59} & \change{47.65} & 47.11 \\
                \textbf{MaSoN}        & \change{38.13 $\pm 0.8$}
                     & \change{91.87 $\pm 1.3$}
                     & 53.88 $\pm 0.6$ \\
             \bottomrule
            \end{tabular}
\end{table}

\section{Ablation Study}
\label{sec:abl}

In this section, we experimentally validate the contributions proposed in this paper. We report the average F1 of five datasets repeated with 5 random seeds for every experiment. The full individual dataset results with additional metrics are in Supplementary G, additional ablations in Supplementary K, and ablation implementation details in Supplementary R.

\begin{table}[!h]
\centering
            \caption{Change generation ablation.}
            \label{tab:sample_gen}
            \begin{tabular}{llc}
            \toprule
            	ID & & \textit{Avg.} F1\\
            \midrule
    \rowcolor{gray!20}\textit{\textbf{Ours}} & Fully latent generation & 50.6{\scriptsize$\pm0.4$}\\
    \textit{Rect. $M_{\text{C}}$} & Random rectangles $M_{\text{C}}$ & 46.9{\scriptsize$\pm0.8$}\\
    \textit{No $M_{\text{C}}$} & No bin. mask for changed & 36.5{\scriptsize$\pm1.7$}\\
    \textit{CycGan} & Irrelevant with CycGAN & 36.9{\scriptsize$\pm0.7$}\\
    \textit{No Dyn.} & No dynamic estimation & 25.3{\scriptsize$\pm1.9$}\\
    \textit{No Irrel.} & No irrelevant changes & 15.6{\scriptsize$\pm0.8$}\\
    \textit{Pix. Gen.} & Pixel noise generation & 12.8{\scriptsize$\pm1.8$}\\
                \bottomrule
                \end{tabular}
\end{table}

\noindent\textbf{Core ablation studies.}
We validate our latent-space change generation strategy through a series of ablations presented in \cref{tab:sample_gen}. Replacing the binarised Perlin mask with a random rectangles mask (\textbf{Rect. $M_{\text{C}}$}) slightly lowers performance, because most changes are non-rectangular and since our approach models them better due to its higher randomness. Removing the binary mask for relevant noise (\textbf{No $M_{\text{C}}$}), where noise is instead applied to half of the features and the ground truth for those is set to "all changed", reduces performance by 14.1 p.p., confirming the benefit of spatially structured perturbations. 

Next, we replace the irrelevant latent noise $\varepsilon_I$ with CycleGAN~\cite{zhu2017cycGan}-transfor\-med images that simulate irrelevant changes in pixel space via a learned transformation (\textbf{CycGan}). This reduces the F1 by 13.7 p.p., highlighting the advantage of modelling irrelevant variation directly in latent space using our approach.

Removing dynamic noise estimation (described in \Cref{sub:changen}) and instead using a fixed noise magnitude reduces performance by 25.3 p.p. (\textbf{No Dyn.}). This demonstrates the importance of data-driven estimation of noise from features that change during training and vary across geographical domains. 
Eliminating irrelevant change noise (\textbf{No Irrel.}, refer to~\Cref{eq:irrel}) decreases the performance by 35.0 p.p. as the model incorrectly learns that unchanged regions correspond to zero feature difference. 

Together, these results show that both mechanisms are essential. Unlike machine learning works from other fields (e.g., anomaly detection with low intra-sample variation - see \Cref{s:rel}), MaSoN \textit{dynamically} estimates two \textit{decoupled} noise components based on feature statistics. In contrast to these other fields, \textit{we show that this is crucial} for bi-temporal remote sensing imagery, where variability is high across geographical regions and within image pairs themselves.

Applying the noise perturbation directly to the pixels of input images (\textbf{Pix. Gen.}) leads to the poorest result (12.8 F1), showing that raw pixel-level noise lacks the semantic robustness and is not as nicely behaved as latent feature space.

\noindent\textbf{Quantile Sampling Dimension.} MaSoN calculates the quantiles per channel in batch (\Cref{sub:changen}). It could, however, use the same value across all channels in a batch, or separate values for each sample, or even separate values for each channel within a sample. The results of such strategies are presented in \Cref{tab:q_dim}. We hypothesise that our strategy of calculating statistics per channel in batch yields the best performance due to generalisation, since the scope of sampling is neither too general (per-batch) nor too specific (per-channel).

\begin{table*}[!t]
    \begin{minipage}[t]{.32\linewidth}
        \caption{Noise sampling dimension.}
        \label{tab:q_dim}
            \centering
            \begin{tabular}{lcccccc}
            \toprule
            	~& \textit{Avg.} F1\\
            \midrule
        \rowcolor{gray!20}Per-channel in batch \scriptsize{(Ours)} & 50.6{\scriptsize$\pm0.4$}\\
        Per-channel in sample & 29.6{\scriptsize$\pm10.7$}\\
        Per-sample & 44.3{\scriptsize$\pm0.8$}\\
        Per-batch & 32.9{\scriptsize$\pm1.7$}\\
            \bottomrule
            \end{tabular}
    \end{minipage}
    \hfill
    \begin{minipage}[t]{.32\linewidth}
        \caption{Backbone ablation.}
        \label{tab:backbone}
        \setlength{\tabcolsep}{4pt}
        \centering
        \begin{tabular}{lcccccc}
        \toprule
        	Architecture & \textit{Avg.} F1\\
        \midrule
        \rowcolor{gray!20}ViT-L (DINOv3) \scriptsize{(Ours)} & 50.0{\scriptsize$\pm0.7$}\\
        ViT-L (DINOv2) & 47.4{\scriptsize$\pm1.5$}\\
        ViT-B (DINOv3) & 45.4{\scriptsize$\pm2.3$} \\
        Swin-T (ADE20k) & 41.7{\scriptsize$\pm1.2$}\\
        ResNet50 (ADE20k) & 34.1{\scriptsize$\pm0.5$}\\
        \bottomrule
        \end{tabular}
    \end{minipage}
    \hfill
    \begin{minipage}[t]{0.32\linewidth}
        \caption{Target layers for noise.}
        \label{tab:layer}
        \centering
        \setlength{\tabcolsep}{2pt}
                \begin{tabular}{lccccc}
            \toprule
                L. 7 & L. 11 & L. 15 & L. 23 & \textit{Avg.} F1\\
                \midrule
        \rowcolor{gray!20}$\checkmark$&$\checkmark$&$\checkmark$&$\checkmark$ & 50.6{\scriptsize$\pm0.4$}\\
        $\checkmark$&$\checkmark$&$\checkmark$&~ & 48.6{\scriptsize$\pm0.3$}\\
        $\checkmark$&$\checkmark$&~&$\checkmark$ & 44.6{\scriptsize$\pm1.6$}\\
        $\checkmark$&~&$\checkmark$&$\checkmark$ & 46.2{\scriptsize$\pm1.6$}\\
        ~&$\checkmark$&$\checkmark$&$\checkmark$ & 48.7{\scriptsize$\pm1.2$}\\
        
            \bottomrule
            \end{tabular}
            \end{minipage}
\end{table*}

\noindent\textbf{Backbone Ablation.} MaSoN's performance does not stem solely from DINOv3~\cite{simeoni2025dinov3}, but it does achieve better performance with a stronger encoder, as shown in \Cref{tab:backbone}. If we replace it with DINOv2~\cite{oquab2024dinov2}, and some weaker backbones like Swin~\cite{liu2021swin} and ResNet50~\cite{he2016deep}, it still remains effective and performs better or comparable to existing approaches. We also opt for DINOv3 as it is also used by two of the best-performing competing methods.

\noindent\textbf{Layer Noise Importance.} To verify the importance of injecting noise into each layer, we conducted an experiment where we excluded noise injection for individual layers. The results in Table~\ref{tab:layer} show that not applying noise to the middle 15th layer brings the largest drop in performance (6 p.\ p.). Nevertheless, all four layers play an important role in the overall performance.

\begin{figure}[!h]
    \centering
    \includegraphics[width=1\linewidth]{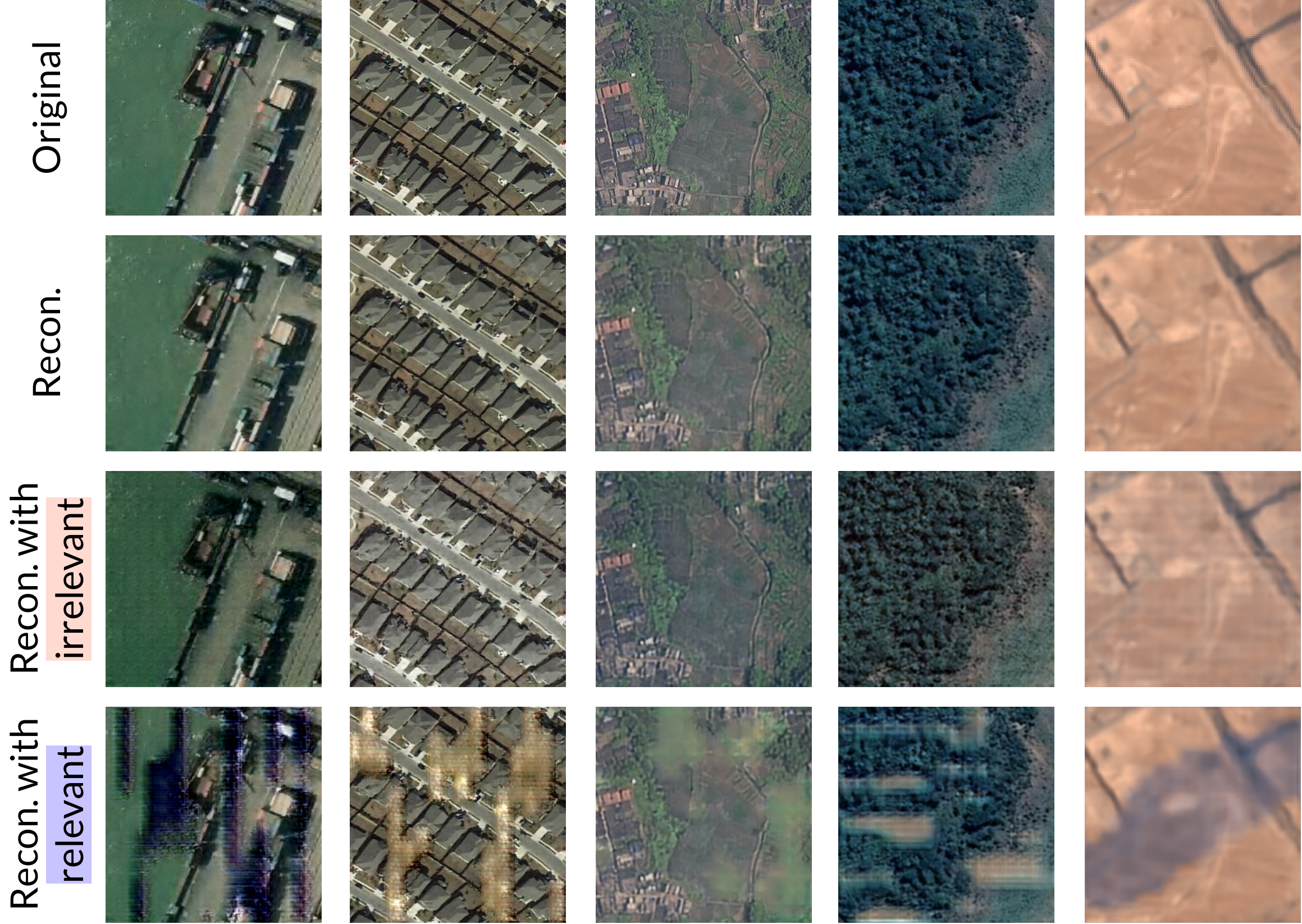}
    \caption{First row contains original input image, second row reconstructions from features, third row reconstructions with \colorbox{irchange!30}{irrelevant} noise added to features, and the final row reconstructions with \colorbox{rchange!30}{relevant} noise added to selected regions (as explained in \cref{sub:changen}). The addition of smaller \colorbox{irchange!30}{irrelevant} noise in the latent space is reflected in a minor colour and texture variation. The addition of \colorbox{rchange!30}{relevant} noise is expressed as a significant but coherent change, even in reconstructed pixel space.}
    \label{fig:recon}
\end{figure}

\noindent\textbf{Feature Analysis.}
We analyse how the changes, synthesised by adding noise in the latent space, are expressed when reconstructed to pixel space. We train a simple UNet~\cite{ronneberger2015u} decoder to reconstruct the input image from the features (see Supplementary E for more details). The results shown in \Cref{fig:recon} display both the reconstructions without noise and with the addition of either irrelevant or relevant noise, where noise is generated as explained in \Cref{sub:changen}. The changes synthesised using our methodology in the latent space also represent a meaningful shift in the pixel space. This further demonstrates how a well-behaved structure of latent space enables a mechanism as simple as Gaussian noise to produce changes that approximate real changes. Based on this and the insights from the analysis in \Cref{sub:feat_analysis}, we believe that latent-space generation has significant potential for future work.

\begin{figure}
    \centering
    \includegraphics[width=0.8\linewidth]{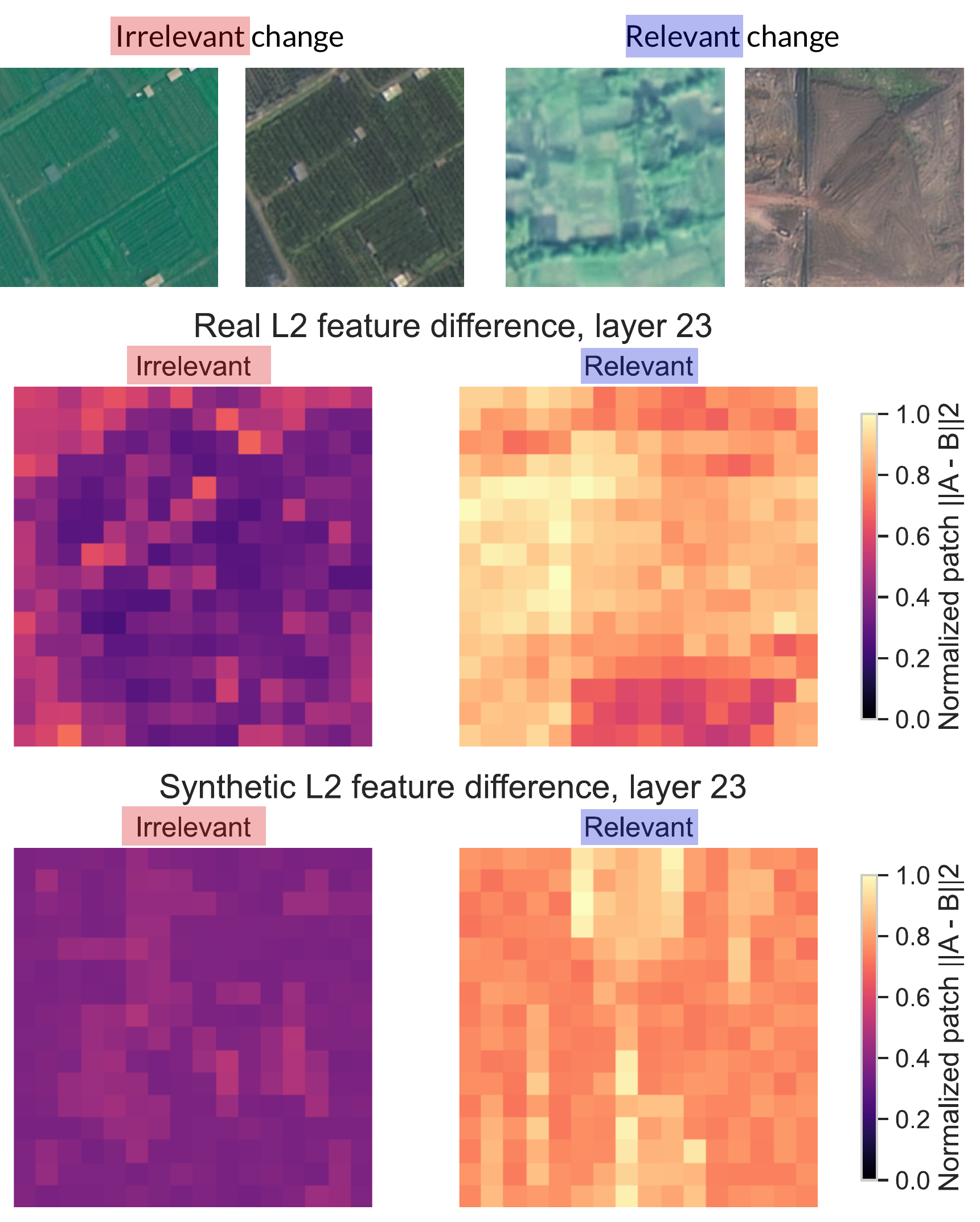}
    \caption{\change{First row contains visual examples being considered, second row shows real feature difference L2 map, and the last row illustrates the feature difference L2 map with synthetically generated changes using our approach. Our strategy approximates the true change range, while the spatial distribution is more random (see SubSec~\ref{sub:changen}).}}
    \label{fig:feat_vis}
\end{figure}

\change{We see how the feature difference L2 map is expressed for real changes and synthetic changes in Figure~\ref{fig:feat_vis}. The range of synthetic changes approximates the one in real ones, while the spatial distribution of change is different due to randomness and since each image in a pair is treated as a single temporal (see SubSec~\ref{sub:changen}).}

\noindent\textbf{Limitations and Future Work.} 
MaSoN could be viewed as a self‑supervised approach within the broader field of unsupervised change detection~\cite{ding2025labelEffsurvey}. Consequently, its limitation is the need for light finetuning rather than full training-free operation. Given its stronger accuracy than zero-shot alternatives and its short training time, we consider this a reasonable trade-off. Moreover, even compared to current zero-shot methods that are prohibitively slow at inference (e.g., DynamicEarth processes only 0.4 images/s), MaSoN is more time-efficient (including fine-tuning) for data collections larger than roughly 170 images. We believe an important next step for unsupervised methods is to reduce end-to-end latency by accelerating latent change synthesis and to improve inference throughput. \change{Another limitation is the heterogeneous change detection and very subtle changes, where MaSoN explicitly relies on alignment and discriminability of features in encoders' latent space, which is not the case in some complex scenarios. Cross-modal alignment of the encoder thus remains an important direction of future work.} Additional discussion and societal impact are provided in Supplementary A.

\section{Conclusion}
We propose MaSoN, a framework for unsupervised change detection (UCD) that addresses the limited generalisation of existing approaches, which rely on priors from foundation models in a training-free manner or train on changes generated in pixel space. MaSoN trains on synthetic changes generated via Gaussian noise injected directly into the latent feature space, where the noise is dynamically estimated per-dataset using the data’s own statistics. This enables highly diverse, data-driven change generation without external data or auxiliary generative models. MaSoN yields substantial improvements, outperforming prior unsupervised state-of-the-art by 14.1 percentage points on average across five diverse benchmarks. MaSoN also extends to multispectral and SAR modalities via a simple encoder swap, achieving great results.
Beyond UCD, this work establishes latent-space synthesis as a viable and effective direction for unsupervised learning in general. The idea holds promise for sample generation and representation learning in low-data regimes, offering a path toward more general and data-efficient vision models in remote sensing as well as other domains.

\section*{Acknowledgments}
This work was in part supported by the ARIS research projects J2-60045 (RoDEO) and GC-0006 (GeoAI), research programme P2-0214, and the supercomputing network SLING (ARNES, EuroHPC Vega). 

\bibliography{main}
\bibliographystyle{IEEEtran}

\input{supp}

\end{document}

%% file: supp.tex
\clearpage 
\setcounter{page}{1} 
\setcounter{footnote}{0} 
\setcounter{table}{0} 
\setcounter{figure}{0} 
\setcounter{equation}{0}  

\maketitlesupplementary 
\phantomsection 
\renewcommand*{\theHsection}{appendix.\arabic{section}} 
\appendix  
\crefalias{section}{appendix} 
\crefalias{subsection}{appendix}

In this Appendix, we provide extensive additional details and supporting information that extend beyond the scope of the main manuscript. The Appendix is organised as follows: 
\begin{itemize}
    \item \textbf{Limitations and Societal Impact} in Section~\ref{a:limit_social}.
    \item \textbf{Extended dataset details} in Section~\ref{a:data}.
    \item \textbf{Extended feature analysis} with per-dataset results and noised feature pixel-space reconstruction in Section~\ref{a:feat}.
    \item \textbf{Theoretical justification of latent Gaussian noise modelling} in Section~\ref{a:maxent}
    \item \textbf{Main results with additional metrics and standard deviation} in Section~\ref{a:ext_res}.
    \item \textbf{Additional qualitative results}, including failure cases in Section~\ref{a:qual}.
    \item \textbf{Supervised results} in Section~\ref{a:sup}
    \item \textbf{Computational efficiency} protocol, extended results, and discussion on noise generation overhead in Section~\ref{a:comp}.
    \item \textbf{Additional ablation studies}, including multispectral and SAR data, batch size effect, statistics sampling, and noise distributions in Section~\ref{a:add_abl}.
    \item \textbf{Extended implementation details} for our model, related works, and ablation studies in Section~\ref{a:ext_impl}.
\end{itemize}

\section{Limitations and Societal Impact}
\label{a:limit_social}

\subsection{Limitations}

MaSoN can be regarded as a self‑supervised approach within the broader UCD paradigm~\cite{ding2025labelEffsurvey}, and this choice introduces some practical limitations. The most central constraint is its reliance on light finetuning rather than operating in a fully zero‑shot setting. Although the required adaptation is minimal, the need for downstream training introduces overhead in scenarios where compute resources are limited and rapid deployment is essential. This stands in contrast to fully zero‑shot methods, which offer maximum flexibility and immediacy but typically at the cost of lower performance. However, compared to current zero‑shot methods that are prohibitively slow (for example, DynamicEarth processes only 0.4 images per second), \textit{MaSoN becomes more time‑efficient overall} (including finetuning) for any \textit{dataset larger than roughly 170 images.}

Despite these constraints, the observed performance gains relative to zero‑sh\-ot approaches, as well as MaSoN’s short and data‑efficient training process, make this training trade‑off reasonable in many practical applications. Nonetheless, reducing or eliminating the finetuning step remains an important direction for future work. Potential improvements include exploring hybrid zero‑/few‑shot mechanisms, more efficient learning objectives, or techniques that enable stron\-ger generalisation from the base representation without task‑specific updates.

Finally, while our empirical results show consistent improvements across evaluated benchmarks, MaSoN’s generality may be constrained when applied to tasks that diverge significantly from the structure or distribution of the data used during its training. Understanding such failure modes and systematically characterising transferability remain open challenges.

\subsection{Societal Impact}

An unsupervised change detection model enables efficient monitoring of environmental and human-driven changes. By eliminating the need for labelled data, our method enables timely and cost-efficient monitoring of a wide range of environmental and human-driven changes, including deforestation, urban expansion, agricultural shifts, and the aftermath of natural disasters. This can be particularly valuable in regions where annotated data is scarce or in applications requiring rapid deployment, such as disaster response or early warning systems.

By avoiding reliance on external auxiliary datasets and operating in an end-to-end manner, MaSoN is suitable for real-world deployment in resource-constrained settings. However, as with all automated systems, care must be taken to interpret results responsibly and to ensure ethical use, particularly in applications that involve sensitive geographic or political regions. Ensuring transparency and human oversight in downstream use is critical to promoting ethical use. We, however, deeply believe that the advantages of our method greatly outweigh the risks and hope that this work brings us closer to enabling safer living conditions for everyone on the planet.

\section{Dataset Details}
\label{a:data}

\begin{table*}[!h]
\resizebox{\linewidth}{!}{
\setlength{\tabcolsep}{2pt}
    \begin{tabular}{lccccccccc}
        \toprule
 & Acquisition & Resolution & Change Type & Interval & Region & \makecell{Image count \\ train\textbackslash val\textbackslash test} & Patch & \makecell{Changed \\ Pixels} & \makecell{Unchanged \\ Pixels} \\
\midrule
SYSU~\cite{shi2022sysuDSAMnet} & Aerial & 0.5m & \makecell{Buidling, urban,\\groundwork, road,\\vegetation, sea} & 2007-2014 & Hong Kong & \makecell{12000\\4000\\4000} & $256\times256$ & $21.8~\%$ & $78.2~\%$ \\\midrule
LEVIR~\cite{chen2020levirStanet} & \makecell{Google Earth \\ satellite} & 0.5m & Buidling & 2002-2018 & \makecell{20 regions \\ in US} & \makecell{7120\\1024\\2048} & $256\times256$ & $4.7~\%$ & $95.3~\%$ \\\midrule
GVLM~\cite{zhang2023gvlm} & \makecell{Google Earth \\ satellite \\ (SPOT-6)} & 0.59m & Landslide & 2010-2021 & \makecell{17 regions\\ worldwide} & \makecell{4558\\1519\\1519} & $256\times256$ & $6.6~\%$ & $93.4~\%$ \\\midrule
CLCD~\cite{li2022clcdMSCANET} & \makecell{Satellite \\ (Gaofen-2)} & 0.5m-2m & \makecell{Multiple types\\limited to\\croplands} & 2017-2019 & \makecell{Guangdong,\\ China} & \makecell{1440\\480\\480} & $256\times256$ & $7.6~\%$ & $92.4~\%$ \\\midrule
OSCD~\cite{daudt2018urban} & \makecell{Satellite \\ (Sentinel-2)} & 10m & Urban & 2015-2018 & \makecell{24 regions \\ worldwide} & \makecell{827\\-\\385} & $96\times96$ & $3.2~\%$ & $96.8~\%$ \\
\bottomrule
    \end{tabular}
    }
    \caption{Additional details for the datasets used in the paper.}
    \label{atab:data_extend}
\end{table*}

Extended dataset specifications are presented in Table~\ref{atab:data_extend}. The datasets encompass a broad range of change types, including building and urban development, cropland shifts, and natural disaster impacts. They also differ in spatial resolution (GSD), acquisition sensors, and dataset size: ranging from a few hundred to several thousand image pairs. This diversity contributes to the robustness and broader applicability of our findings.

A common challenge in remote sensing change detection is the highly imbalanced nature of the data, as changed pixels typically account for less than $10\%$ of the total. An exception is the SYSU dataset, which shows a higher change ratio due to its coarse annotations over large change regions.

\subsection{Data Implementation Details}
\label{asub:data_impl}

\textbf{Dataset splits} Official train-test splits are used for OSCD, SYSU, CLCD, and LEVIR, while GVLM relies on already available public random splits from Huggingface (courtesy of Weikang Yu) to ensure full reproducibility and fair comparison.

The source for the data used are as follows:
\begin{itemize}
    \item SYSU: \textcolor{weights}{ericyu: SYSU\_CD}
    \item LEVIR: \textcolor{weights}{ericyu: LEVIRCD\_Cropped256}
    \item GVLM: \textcolor{weights}{ericyu: GVLM\_Cropped\_256}
    \item CLCD: \textcolor{weights}{ericyu: CLCD\_Cropped\_256}
    \item OSCD: \textcolor{weights}{Official webpage}. Since this dataset is not prepared, we manually crop the images from a predefined train-test split into correctly shaped patches. This dataset also provides multispectral data, which we will evaluate later in further ablation studies.
\end{itemize}

\section{Extended Feature Analysis}
\label{a:feat}

\subsection{Per-Dataset Feature Differences}

We use a trained model setup as described in Subsection~\ref{a:ext_impl} for supervised learning. The features are then extracted from each image in the test set for every dataset used, and feature differences are computed. This is done for each hierarchical layer in the multi-scale features elementwise. A per-channel histogram of these differences is calculated, and the mean histogram of all channels is visualised.

\begin{figure}[!h]
    \centering
    \includegraphics[width=1\linewidth]{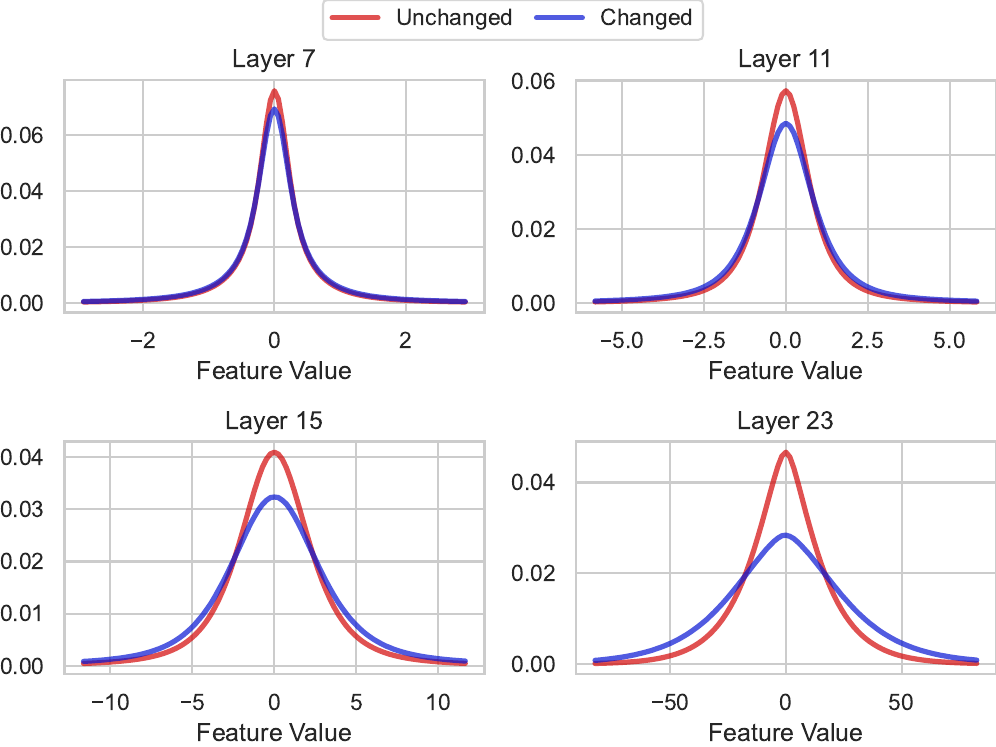}
    \caption{Distributions of feature differences $f_1^{(l)} - f_2^{(l)}$, on the SYSU Dataset.}
    \label{fig:sysu_hist}
\end{figure}

\begin{figure}[!h]
    \centering
    \includegraphics[width=1\linewidth]{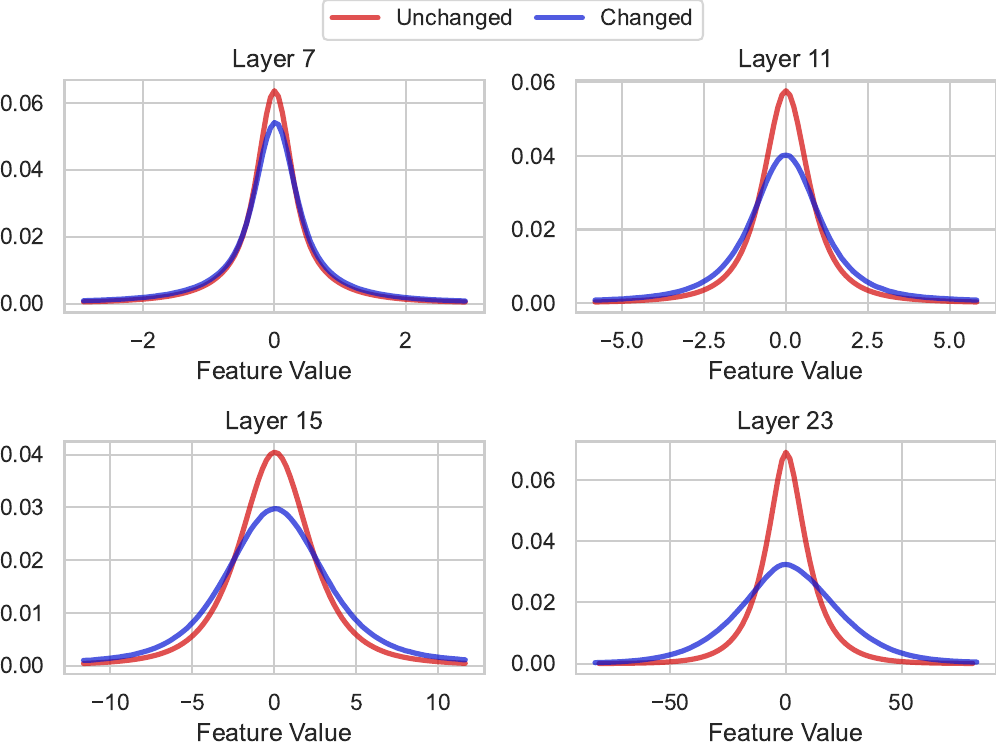}
    \caption{Distributions of feature differences $f_1^{(l)} - f_2^{(l)}$, on the LEVIR Dataset.}
    \label{fig:levir_hist}
\end{figure}

\begin{figure}[!h]
    \centering
    \includegraphics[width=1\linewidth]{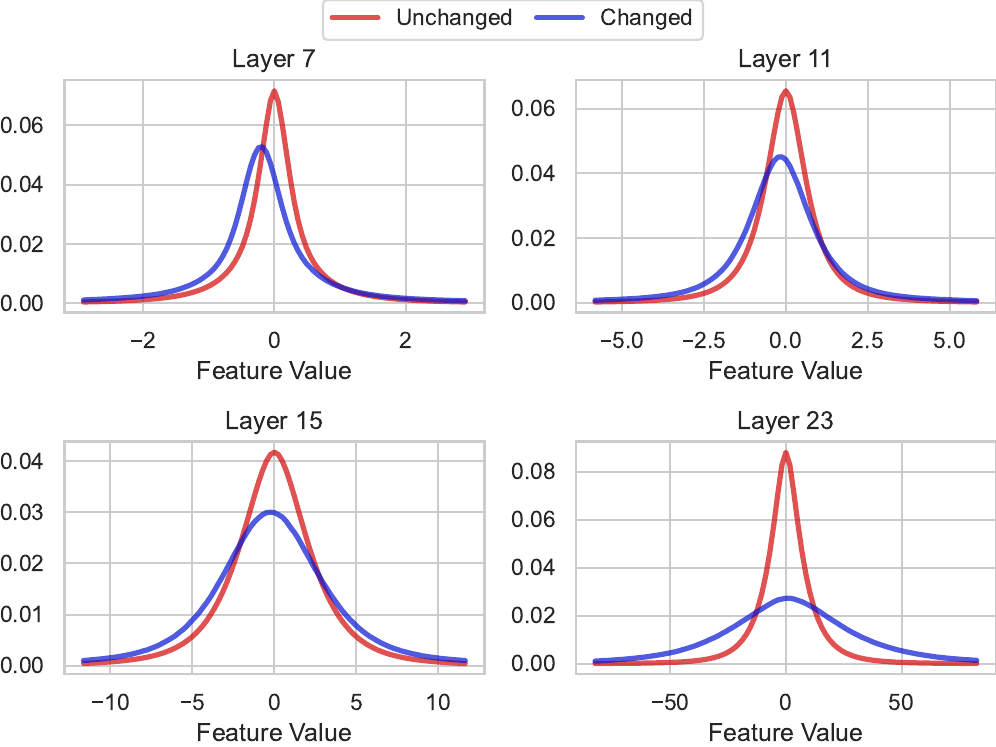}
    \caption{Distributions of feature differences $f_1^{(l)} - f_2^{(l)}$, on the GVLM Dataset.}
    \label{fig:gvlm_hist}
\end{figure}

\begin{figure}[!h]
    \centering
    \includegraphics[width=1\linewidth]{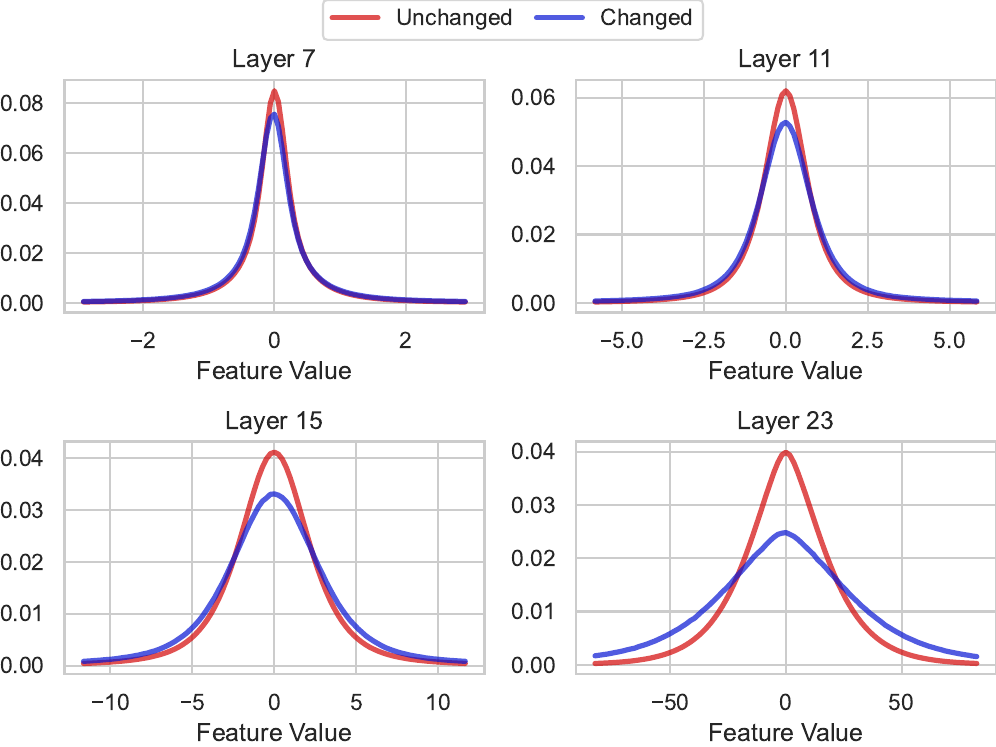}
    \caption{Distributions of feature differences $f_1^{(l)} - f_2^{(l)}$, on the CLCD Dataset.}
    \label{fig:clcd_hist}
\end{figure}

\begin{figure}[!h]
    \centering
    \includegraphics[width=1\linewidth]{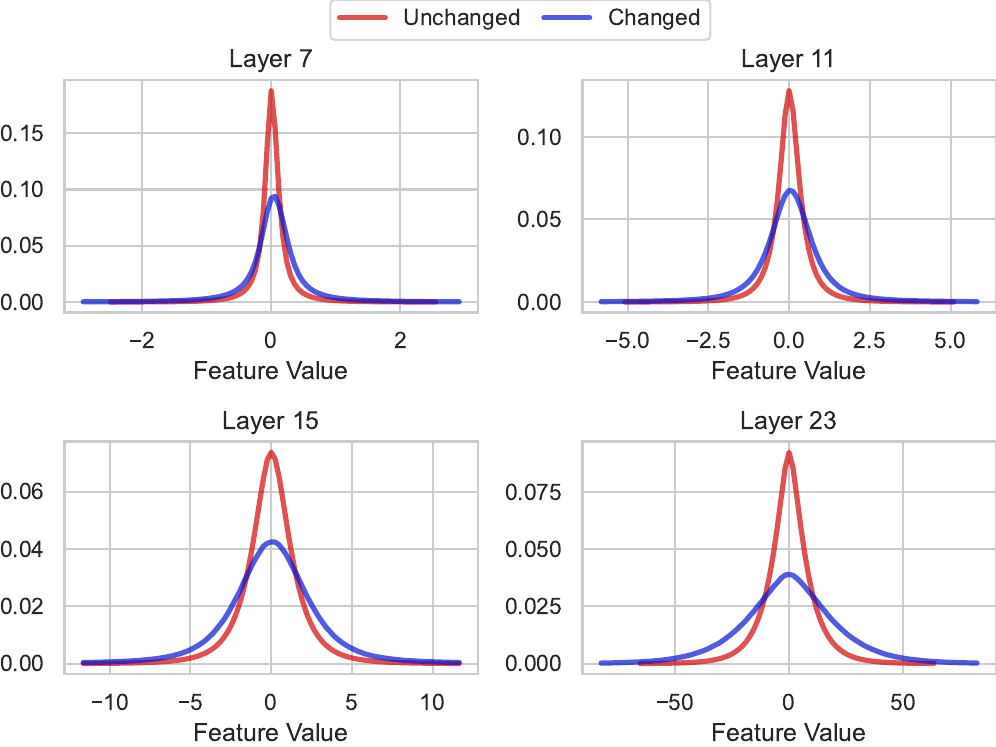}
    \caption{Distributions of feature differences $f_1^{(l)} - f_2^{(l)}$, on the OSCD Dataset.}
    \label{fig:oscd_hist}
\end{figure}

The average histogram over all datasets is reported in the main paper. Here, we also include an individual mean histogram for all the datasets used in our study. They can be seen in Figures~\ref{fig:sysu_hist}-~\ref{fig:oscd_hist}. The features of the changed and unchanged regions exhibit similar Gaussian-like behaviour across all datasets, but the distributions slightly differ for each one.

\subsection{Noised Feature reconstruction}
\label{a:feat_recon}

To assess the semantic impact of our latent space perturbation strategy, we conducted a reconstruction-based visualisation experiment. Starting with the supervised model detailed in Subsection~\ref{a:ext_impl} for supervised learning, we froze the encoder and trained a separate U-Net-like decoder (different from the change mask decoder) that simply reconstructs the original input image from the encoder's feature representations. This reconstruction decoder was trained for 100 epochs using an L2 loss function. 

\begin{figure}[!h]
    \centering
    \includegraphics[width=1\linewidth]{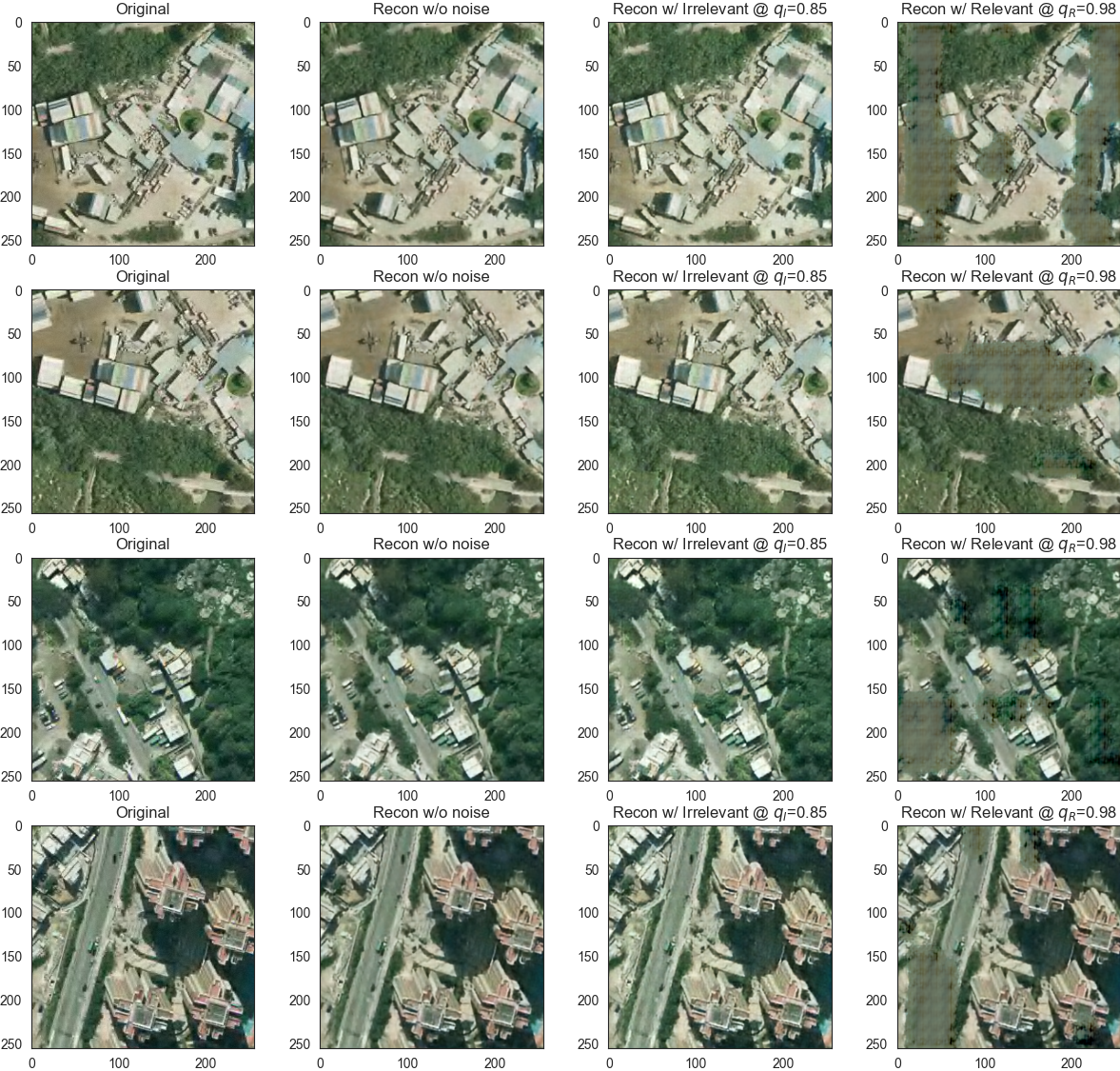}
    \caption{Qualitative examples of feature reconstructions on SYSU Dataset. In the first column, the original image is shown, in the second, the reconstructed image without any noise, in the third, the features with the addition of irrelevant noise and in the fourth, the features with the addition of relevant noise (added only to some sections of the image).}
    \label{fig:recon_sysu}
\end{figure}

\begin{figure}[!h]
    \centering
    \includegraphics[width=1\linewidth]{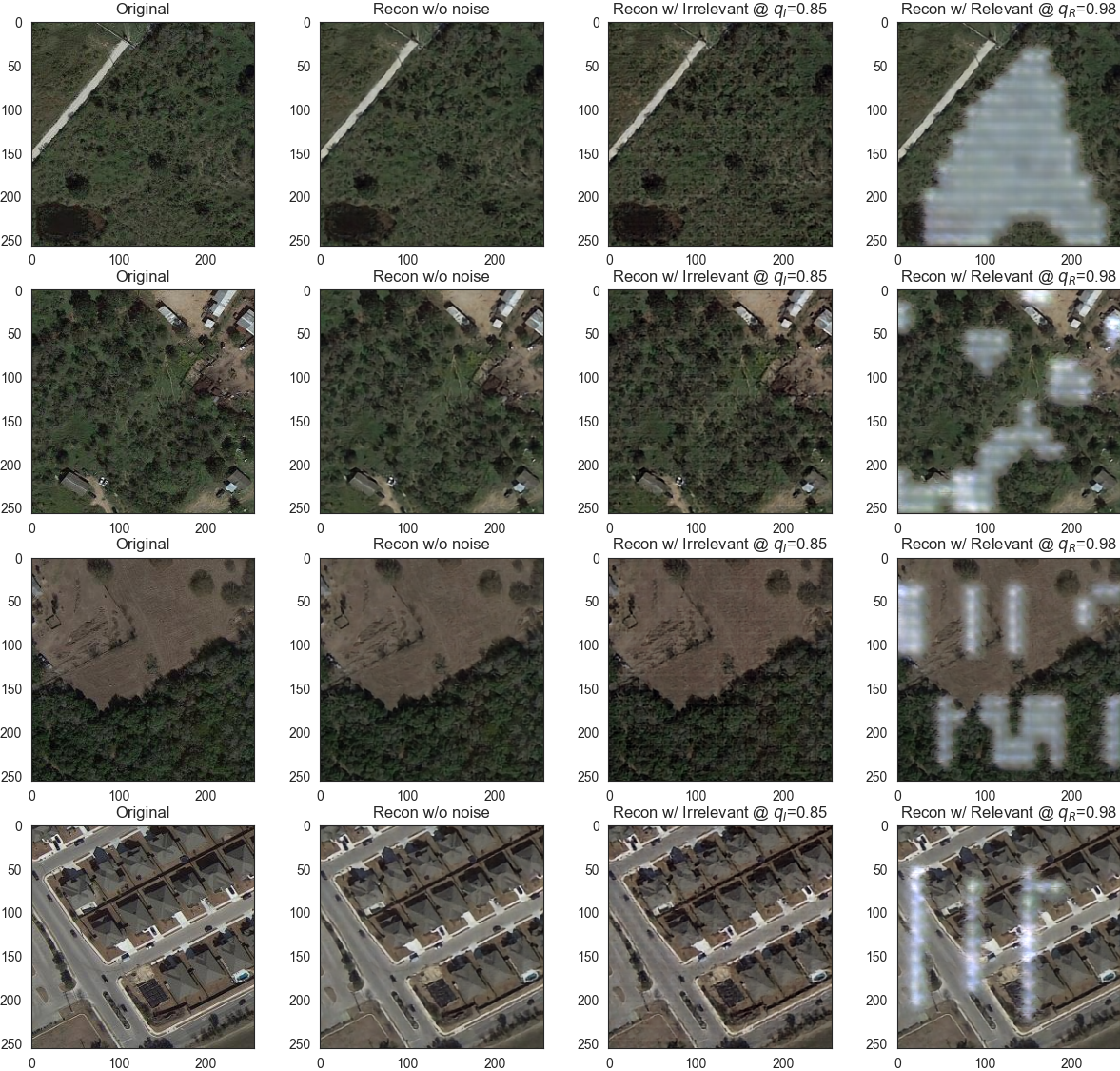}
    \caption{Qualitative examples of feature reconstructions on LEVIR Dataset.}
    \label{fig:recon_levir}
\end{figure}

\begin{figure}[!h]
    \centering
    \includegraphics[width=1\linewidth]{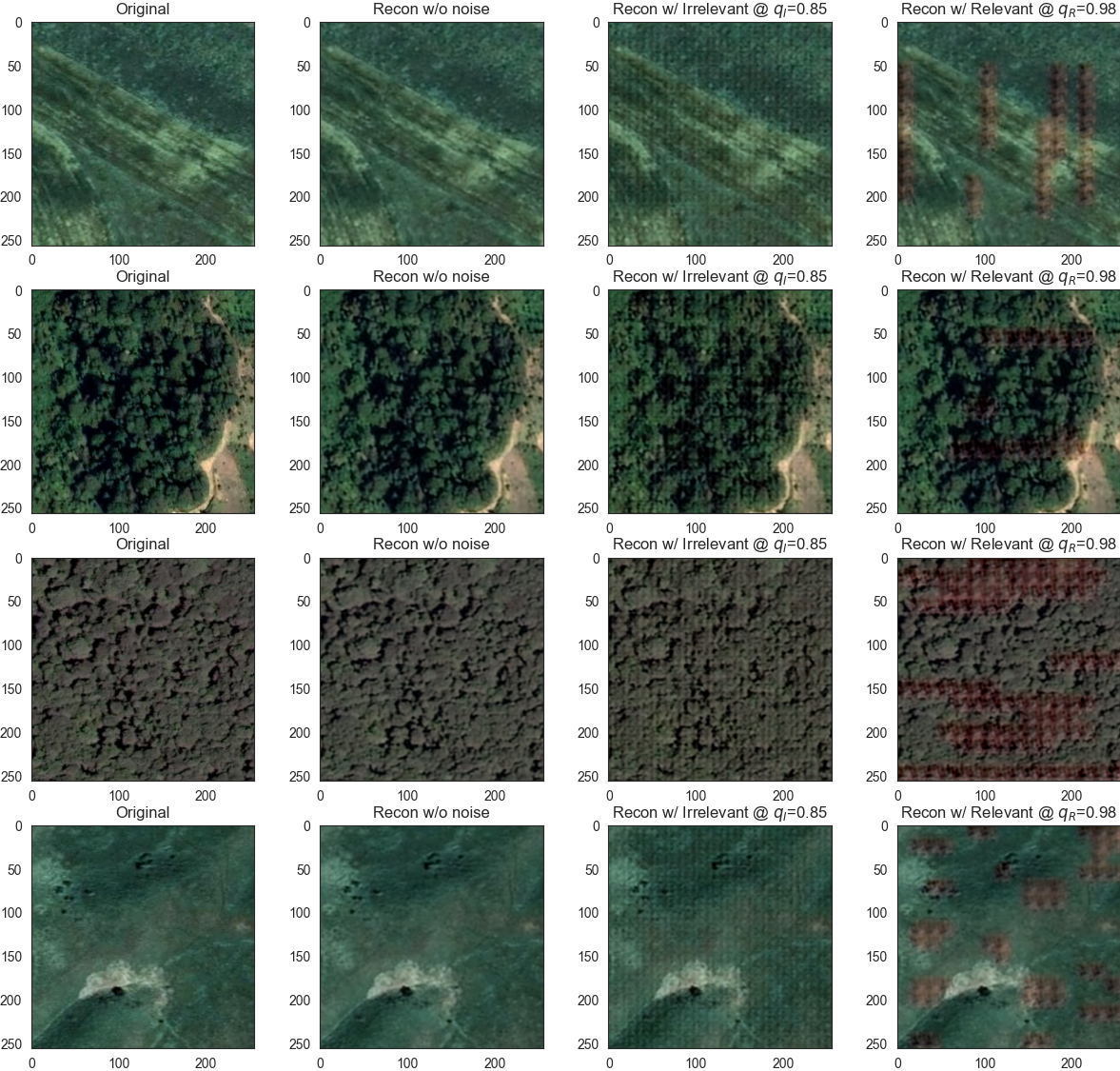}
    \caption{Qualitative examples of feature reconstructions on GVLM Dataset.}
    \label{fig:recon_gvlm}
\end{figure}

\begin{figure}[!h]
    \centering
    \includegraphics[width=1\linewidth]{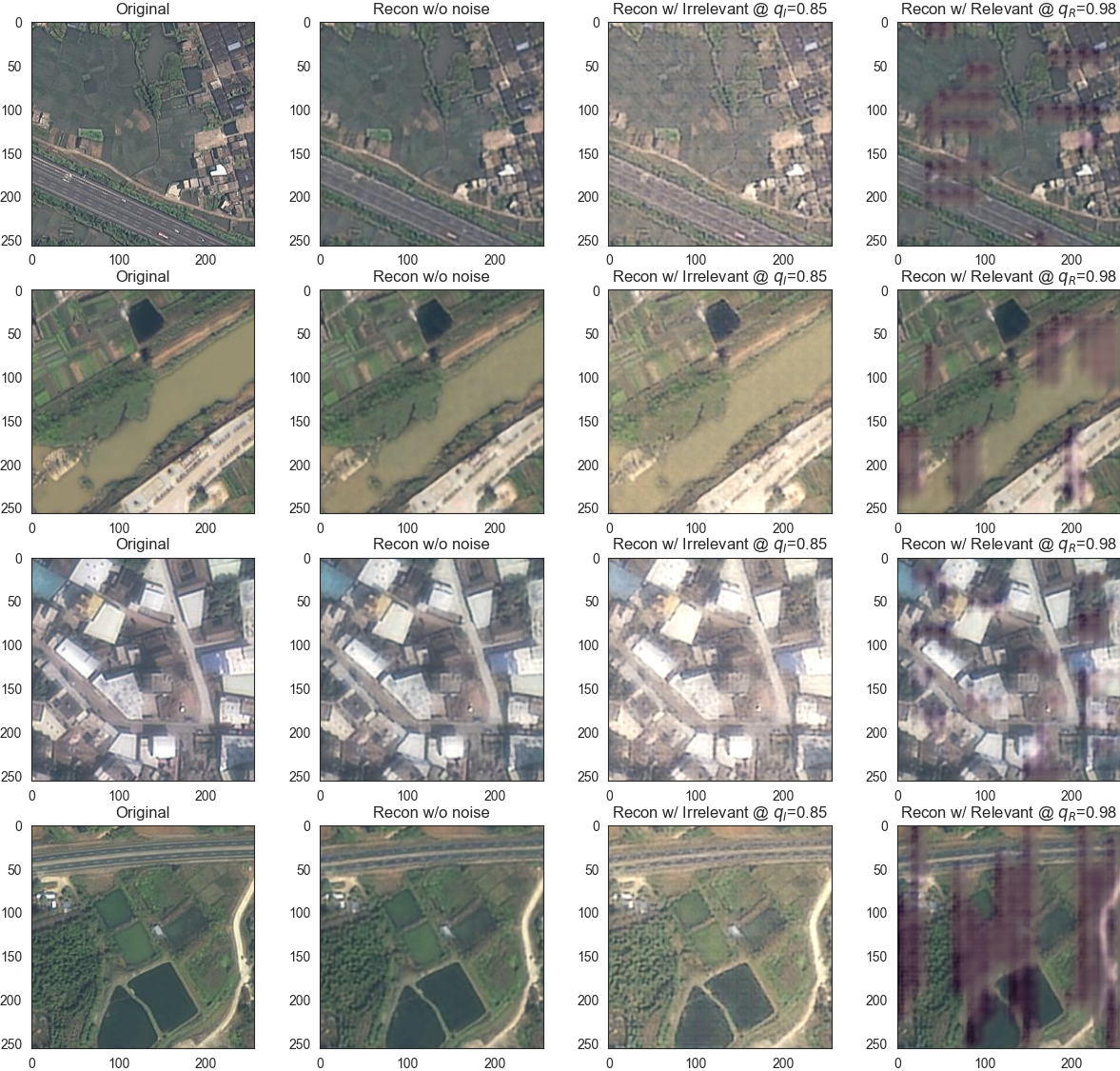}
    \caption{Qualitative examples of feature reconstructions on CLCD Dataset.}
    \label{fig:recon_clcd}
\end{figure}

\begin{figure}[!h]
    \centering
    \includegraphics[width=1\linewidth]{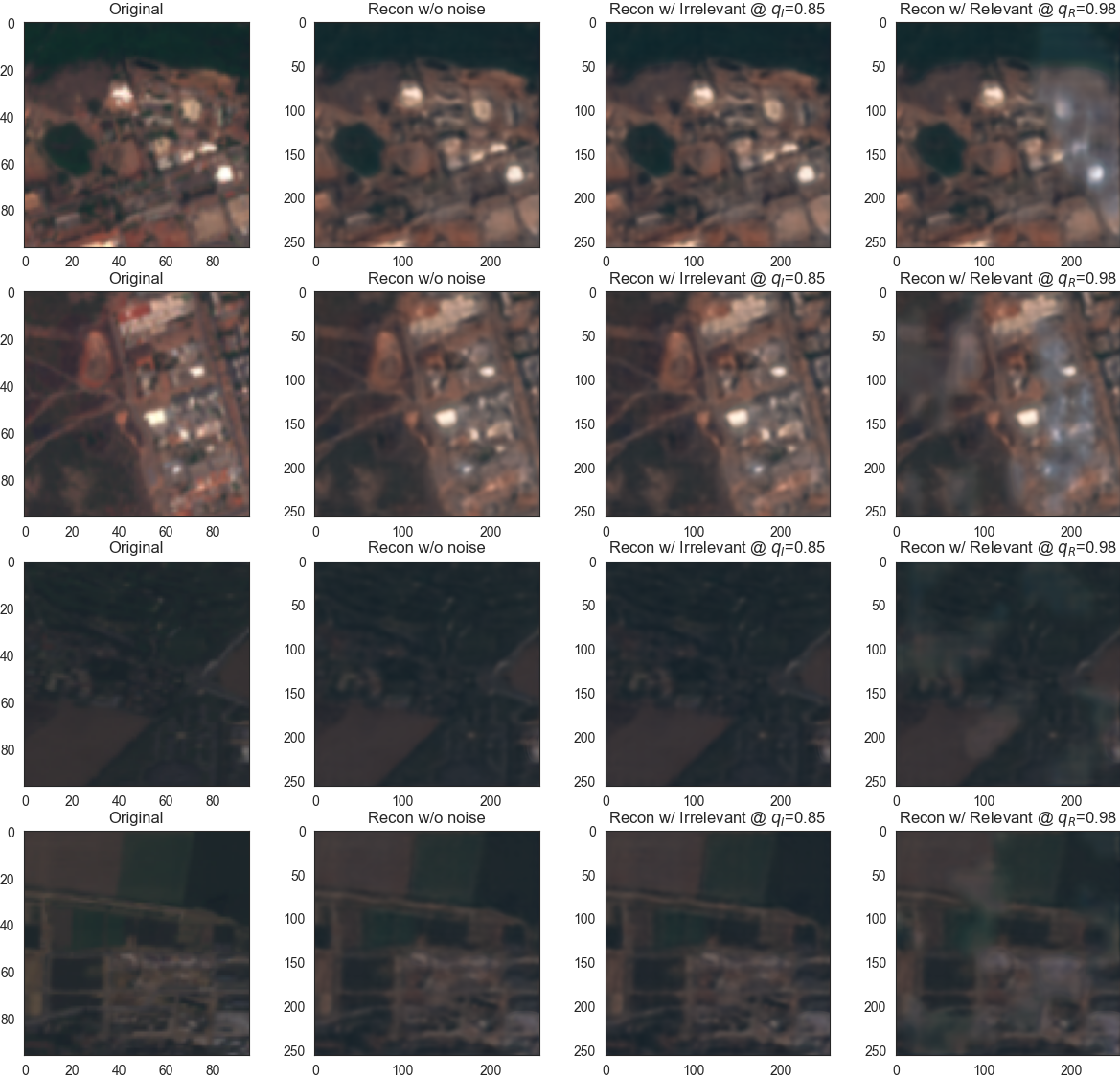}
    \caption{Qualitative examples of feature reconstructions on OSCD Dataset.}
    \label{fig:recon_oscd}
\end{figure}

After training, we extracted features from the original encoder and applied noise sampled identically to our unsupervised change generation strategy. These perturbed features were then passed through the reconstruction decoder to generate image representations. The resulting reconstructions (Figure~\ref{fig:recon_sysu}-~\ref{fig:recon_oscd}) reveal that the injected noise induces plausible and semantically meaningful changes, mimicking real-world variations. 

Added irrelevant noise manifests as a small variation in texture and colour, but the larger relevant noise is reflected as a noticeable change. While not totally realistic, it interestingly resembles changes like the addition of dirt or concrete or a change to grassland in some regions. 

We also notice that a similar type of change occurs across the entire batch, which is a result of our per-channel in-batch sampling strategy. This strengthens our claim that the synthetic changes introduced by our strategy produce meaningful perturbations within the latent space.

\clearpage

\section{Theoretical justification of latent Gaussian noise modelling}
\label{a:maxent}

From our empirical analysis (see \Cref{sub:feat_analysis}), we infer the constraints $\mathbb{E}[\varepsilon]=0$ and that the per-channel variance of relevant changes is larger than that of irrelevant changes, i.e., $\sigma_R^2 > \sigma_I^2$. We further simplify the modelling by injecting noise independently per channel (diagonal covariance). Concretely, we dynamically estimate per-channel scale parameters (e.g., via quantile-based spread estimates) from the data during training and use them to fix the corresponding second-order statistics (Eq.~(2)–(5) of the main paper). The remaining question is then: given only these constraints, what distribution should we assign to $\varepsilon$?

The principle of maximum entropy provides an answer: among all distributions that satisfy the known constraints, the one with the largest entropy is the least biased and most faithful to the current state of knowledge~\cite{jaynes1957maxEnt}.

Among all continuous distributions on $\mathbb{R}$ with fixed mean and variance, the Gaussian satisfies the constraints and achieves the maximum-entropy~\cite{cover2005infoTheo}:
\begin{equation}
    \varepsilon \sim \mathcal{N}(0,\sigma^2),
\end{equation}

Thus, modelling latent irrelevant and relevant changes as channel-wise Gaussian noise does not require assuming that individual patch values or patch differences are Gaussian. Rather, it is the maximum-entropy choice implied by the first- and second-order per-channel statistics we estimate from data. Empirically, we also observe that aggregate feature-difference distributions are approximately Gaussian (see \Cref{sub:feat_analysis}), and MaSoN achieves strong results with this modelling choice, suggesting that per-channel Gaussian noise is sufficient in practice.

This theoretically motivates our design while leaving room for future work exploring 
more structured, constrained, or data-adaptive modelling beyond the maximum-entropy Gaussian baseline.

\section{Results With Complete Metrics and Standard Deviation}
\label{a:ext_res}

\subsection{Complete Unsupervised Results}

In Table~\ref{atab:unsup}, the complete results for the unsupervised setting are presented, including precision, recall, and F1 with the accompanying standard deviation. The results show that MaSoN balances recall and precision well and achieves the best F1. 

\begin{table*}[!ht]
    \centering
    \caption{Unsupervised change detection results across five diverse datasets and their average. We report Precision (Pr.), Recall (Re.), and F1 score. The symbol * indicates methods for which reproducibility with public code was not possible. Methods without standard deviations are deterministic. \textcolor{goldD}{First} and \textcolor{silverD}{second} place results are marked.}
    \label{atab:unsup}
    \resizebox{\linewidth}{!}{
\setlength{\tabcolsep}{2pt}
    \begin{tabular}{lccc|ccc|ccc|ccc|ccc|ccc}
    \toprule
    	\multirow{2}{*}{~}& \multicolumn{3}{c}{SYSU} & \multicolumn{3}{c}{LEVIR} & \multicolumn{3}{c}{GVLM} & \multicolumn{3}{c}{CLCD} & \multicolumn{3}{c}{OSCD} & \multicolumn{3}{c}{\textit{Avg}} \\
& Pr. & Re. & F1& Pr. & Re. & F1& Pr. & Re. & F1& Pr. & Re. & F1& Pr. & Re. & F1& Pr. & Re. & F1\\
 \hline
Pixel Difference & 40.5 & 45.7 & 42.9 & 5.1 & 35.1 & 8.9 & 14.0 & 70.7 & 23.4 & 9.9 & 43.4 & 16.1 & 11.3 & 51.5 & 18.5 & 16.2 & 49.3 & 22.0\\
DCVA~\cite{saha2019unsupervised}\tiny{TGRS19} & 63.0 & 1.3 & 2.5 & 3.6 & 0.1 & 0.2 & \bm2{ 33.4} & 1.7 & 3.2 & 13.4 & 1.7 & 3.1 & \bm2{ 36.6} & 41.6 & \bm2{ 38.9} & 30.0 & 9.3 & 9.6\\
DinoV2-CVA & 36.4 & \bm1{ 81.2} & 50.3 & 8.6 & \bm1{ 93.9} & 15.7 & 10.2 & \bm2{ 81.1} & 18.2 & 13.0 & \bm1{ 87.6} & 22.7 & 10.1 & \bm2{ 81.8} & 18.1 & 15.7 & \bm1{ 85.1} & 25.0\\
DINOv3-CVA & 44.7 & \bm2{ 76.7} & 56.5 & 10.7 & \bm2{ 91.8} & 19.1 & 11.3 & 65.4 & 19.3 & 16.1 & \bm2{ 85.8} & 27.1 & 12.3 & \bm1{ 82.2} & 21.3 & 19.0 & \bm2{ 80.4} & 28.7\\
AnyChange~\cite{zheng2024anychange}\tiny{NeurIPS24} & 48.1 & 44.1 & 46.0 & 15.2 & 77.6 & 25.4 & 17.1 & 26.7 & 20.8 & 19.1 & 43.5 & 26.5 & 30.3 & 22.9 & 26.1 & 26.0 & 43.0 & 29.0\\
Changen2-S9~\cite{zgeng2025changen2}\tiny{TPAMI25} & 40.8 & 49.6 & 44.8 & 11.5 & 73.0 & 19.9 & 11.9 & 48.3 & 19.1 & 11.3 & 25.2 & 15.6 & 10.9 & 0.8 & 1.5 & 17.3 & 39.4 & 20.2\\
CDRL*~\cite{noh2022cdrl}\tiny{CVPRW22} & \meanwithstd{24.8}{0.6} & \meanwithstd{1.5}{0.1} & \meanwithstd{2.9}{0.1} & \meanwithstd{3.8}{0.1} & \meanwithstd{2.1}{0.1} & \meanwithstd{2.7}{0.1} & \meanwithstd{13.5}{0.0} & \meanwithstd{11.6}{0.0} & \meanwithstd{12.5}{0.0} & \meanwithstd{0.4}{0.2} & \meanwithstd{0.0}{0.0} & \meanwithstd{0.1}{0.0} & \meanwithstd{13.6}{1.5} & \meanwithstd{0.2}{0.0} & \meanwithstd{0.4}{0.0} & \meanwithstd{11.2}{0.2} & \meanwithstd{3.1}{0.0} & \meanwithstd{3.7}{0.0}\\
CDRL-SA*~\cite{noh2024cdrlSA}\tiny{RSL24} & \meanwithstd{58.4}{6.9} & \meanwithstd{1.3}{0.3} & \meanwithstd{2.6}{0.6} & \meanwithstd{19.2}{2.3} & \meanwithstd{3.0}{0.8} & \meanwithstd{5.1}{1.1} & \meanwithstd{15.2}{9.7} & \meanwithstd{0.2}{0.1} & \meanwithstd{0.4}{0.3} & \meanwithstd{18.8}{11.5} & \meanwithstd{0.4}{0.3} & \meanwithstd{0.7}{0.6} & \meanwithstd{30.2}{10.5} & \meanwithstd{0.3}{0.2} & \meanwithstd{0.5}{0.4} & \meanwithstd{28.4}{4.2} & \meanwithstd{1.0}{0.2} & \meanwithstd{1.9}{0.3}\\
I3PE~\cite{chen2023_i3pe}\tiny{ISPRS23} & \meanwithstd{31.0}{1.9} & \meanwithstd{38.0}{9.7} & \meanwithstd{33.7}{4.2} & \meanwithstd{10.7}{0.3} & \meanwithstd{38.7}{3.7} & \meanwithstd{16.8}{0.6} & \meanwithstd{19.2}{2.6} & \meanwithstd{23.2}{8.0} & \meanwithstd{20.3}{3.1} & \meanwithstd{7.9}{1.0} & \meanwithstd{21.3}{5.4} & \meanwithstd{11.5}{1.8} & \meanwithstd{9.2}{1.3} & \meanwithstd{2.8}{3.8} & \meanwithstd{3.6}{3.8} & \meanwithstd{15.6}{0.3} & \meanwithstd{24.8}{3.1} & \meanwithstd{17.2}{1.6}\\
HySCDG~\cite{benidir2025hyscdg}\tiny{CVPR25} & 44.3 & 65.0 & 52.7 & 8.4 & 90.3 & 15.4 & 9.6 & \bm1{ 91.6} & 17.3 & 15.1 & 69.5 & 24.9 & \bm1{ 68.5} & 9.8 & 17.2 & 29.2 & 65.2 & 25.5\\
SCM~\cite{tan2024scm}\tiny{IGARSS24} & 37.2 & 19.7 & 25.7 & 19.5 & 45.3 & 27.3 & 23.5 & 34.7 & 28.0 & 15.8 & 23.5 & 18.9 & 11.9 & 27.4 & 16.6 & 21.6 & 30.1 & 23.3\\
DynEarth~\cite{li2025dynearth}\tiny{AAAI26} & 52.9 & 59.9 & 56.2 & \bm2{ 33.7} & 73.8 & \bm2{ 46.2} & 11.6 & 29.0 & 16.6 & 22.0 & 45.0 & 29.5 & 33.8 & 21.9 & 26.6 & 30.8 & 45.9 & 35.0\\
DynE. (DINOv3)~\cite{li2025dynearth}\tiny{AAAI26} & 70.2 & 52.8 & \bm2{ 60.3} & \bm1{ 37.3} & 73.2 & \bm1{ 49.5} & 14.8 & 27.7 & 19.3 & 28.3 & 41.6 & 33.7 & 23.9 & 14.7 & 18.2 & 34.9 & 42.0 & 36.2\\
S2C (DINOv3)~\cite{ding2025s2c}\tiny{AAAI26} & \meanwithstd{\bm1{73.3}}{2.8} & \meanwithstd{25.3}{2.0} & \meanwithstd{37.6}{2.5} & \meanwithstd{21.7}{2.2} & \meanwithstd{60.8}{5.9} & \meanwithstd{32.0}{3.1} & \meanwithstd{33.1}{4.6} & \meanwithstd{45.5}{4.1} & \meanwithstd{\bm2{38.1}}{3.2} & \meanwithstd{\bm1{50.6}}{2.0} & \meanwithstd{54.0}{3.9} & \meanwithstd{\bm2{52.2}}{2.8} & \meanwithstd{12.7}{8.1} & \meanwithstd{3.2}{2.3} & \meanwithstd{5.1}{3.6} & \meanwithstd{\bm2{41.4}}{4.2} & \meanwithstd{42.1}{5.9} & \meanwithstd{\bm2{36.5}}{4.5}\\
\textbf{Ours} & \meanwithstd{\bm2{72.3}}{1.2} & \meanwithstd{52.3}{3.7} & \meanwithstd{\bm1{60.6}}{2.2} & \meanwithstd{25.9}{1.2} & \meanwithstd{78.5}{4.2} & \meanwithstd{38.9}{1.2} & \meanwithstd{\bm1{43.2}}{4.9} & \meanwithstd{78.5}{6.1} & \meanwithstd{\bm1{55.4}}{2.9} & \meanwithstd{\bm2{46.3}}{1.7} & \meanwithstd{66.4}{6.8} & \meanwithstd{\bm1{54.3}}{1.5} & \meanwithstd{34.2}{2.1} & \meanwithstd{60.3}{2.8} & \meanwithstd{\bm1{43.5}}{1.0} & \meanwithstd{\bm1{44.4}}{1.9} & \meanwithstd{67.2}{3.8} & \meanwithstd{\bm1{50.6}}{0.4}\\

    \bottomrule
    \end{tabular}
    }
\end{table*}

\subsection{Complete Ablation Results}
\label{asub:complete_abl}

We provide detailed per-dataset results and additional metrics for the ablation studies discussed in the main paper. Table~\ref{atab:gen} presents the complete results for different change generation strategies. Table~\ref{atab:backbone} compares different encoder architectures. Table~\ref{atab:layer} evaluates the impact of injecting noise at different feature layers, and Table~\ref{atab:qdim} analyses the effect of varying the noise sampling dimension.

\begin{table*}[!h]
    \centering
        \caption{Different noise generation techniques}
    \label{atab:gen}
    \resizebox{\linewidth}{!}{
\setlength{\tabcolsep}{2pt}
    \begin{tabular}{lccc|ccc|ccc|ccc|ccc|ccc}
    \toprule
    	\multirow{2}{*}{~}& \multicolumn{3}{c}{SYSU} & \multicolumn{3}{c}{LEVIR} & \multicolumn{3}{c}{GVLM} & \multicolumn{3}{c}{CLCD} & \multicolumn{3}{c}{OSCD} & \multicolumn{3}{c}{\textit{Avg.}} \\
& Pr. & Re. & F1& Pr. & Re. & F1& Pr. & Re. & F1& Pr. & Re. & F1& Pr. & Re. & F1& Pr. & Re. & F1\\
 \hline
Fully latent space generation \scriptsize{(Ours)} & \meanwithstd{72.3}{1.2} & \meanwithstd{52.3}{3.7} & \meanwithstd{60.6}{2.2} & \meanwithstd{25.9}{1.2} & \meanwithstd{78.5}{4.2} & \meanwithstd{38.9}{1.2} & \meanwithstd{43.2}{4.9} & \meanwithstd{78.5}{6.1} & \meanwithstd{55.4}{2.9} & \meanwithstd{46.3}{1.7} & \meanwithstd{66.4}{6.8} & \meanwithstd{54.3}{1.5} & \meanwithstd{34.2}{2.1} & \meanwithstd{60.3}{2.8} & \meanwithstd{43.5}{1.0} & \meanwithstd{44.4}{1.9} & \meanwithstd{67.2}{3.8} & \meanwithstd{50.6}{0.4}\\
Random rectangles mask & \meanwithstd{77.0}{1.2} & \meanwithstd{39.9}{3.5} & \meanwithstd{52.5}{2.7} & \meanwithstd{25.9}{0.5} & \meanwithstd{65.1}{6.8} & \meanwithstd{36.9}{1.3} & \meanwithstd{47.9}{3.1} & \meanwithstd{63.3}{4.8} & \meanwithstd{54.3}{0.8} & \meanwithstd{50.1}{2.2} & \meanwithstd{54.9}{5.5} & \meanwithstd{52.2}{1.7} & \meanwithstd{38.5}{2.4} & \meanwithstd{39.3}{2.8} & \meanwithstd{38.8}{0.4} & \meanwithstd{47.9}{1.8} & \meanwithstd{52.5}{3.9} & \meanwithstd{46.9}{0.8}\\

No Perlin mask for changed & \meanwithstd{88.5}{0.7} & \meanwithstd{20.1}{2.6} & \meanwithstd{32.7}{3.4} & \meanwithstd{26.7}{0.9} & \meanwithstd{55.9}{5.6} & \meanwithstd{36.1}{1.2} & \meanwithstd{58.0}{3.5} & \meanwithstd{51.2}{4.8} & \meanwithstd{54.2}{2.6} & \meanwithstd{60.0}{1.3} & \meanwithstd{37.5}{3.2} & \meanwithstd{46.1}{2.1} & \meanwithstd{44.0}{2.2} & \meanwithstd{8.0}{1.4} & \meanwithstd{13.5}{1.9} & \meanwithstd{55.5}{1.4} & \meanwithstd{34.6}{2.8} & \meanwithstd{36.5}{1.7}\\
Irrelevant changes with CycGAN & \meanwithstd{62.6}{1.7} & \meanwithstd{46.5}{3.3} & \meanwithstd{53.3}{1.6} & \meanwithstd{13.9}{1.0} & \meanwithstd{95.0}{1.7} & \meanwithstd{24.3}{1.5} & \meanwithstd{20.4}{2.4} & \meanwithstd{56.7}{4.1} & \meanwithstd{29.9}{2.2} & \meanwithstd{26.2}{1.2} & \meanwithstd{63.0}{2.1} & \meanwithstd{37.0}{1.0} & \meanwithstd{44.8}{2.6} & \meanwithstd{36.7}{4.4} & \meanwithstd{40.1}{1.9} & \meanwithstd{33.6}{1.0} & \meanwithstd{59.6}{1.8} & \meanwithstd{36.9}{0.7}\\
No dynamic calibration & \meanwithstd{28.3}{1.8} & \meanwithstd{95.9}{2.2} & \meanwithstd{43.7}{1.9} & \meanwithstd{6.0}{0.4} & \meanwithstd{99.5}{0.4} & \meanwithstd{11.3}{0.7} & \meanwithstd{8.8}{1.0} & \meanwithstd{96.4}{2.5} & \meanwithstd{16.1}{1.6} & \meanwithstd{9.4}{0.8} & \meanwithstd{97.5}{1.7} & \meanwithstd{17.1}{1.2} & \meanwithstd{26.1}{4.3} & \meanwithstd{74.3}{7.4} & \meanwithstd{38.3}{3.9} & \meanwithstd{15.7}{1.6} & \meanwithstd{92.7}{2.7} & \meanwithstd{25.3}{1.9}\\
Pixel space generation & \meanwithstd{71.1}{2.9} & \meanwithstd{1.7}{1.1} & \meanwithstd{3.3}{2.0} & \meanwithstd{21.6}{0.6} & \meanwithstd{3.3}{2.0} & \meanwithstd{5.5}{2.9} & \meanwithstd{11.4}{0.7} & \meanwithstd{58.2}{13.9} & \meanwithstd{18.8}{0.8} & \meanwithstd{41.1}{1.9} & \meanwithstd{5.0}{1.8} & \meanwithstd{8.8}{2.9} & \meanwithstd{40.6}{5.0} & \meanwithstd{21.6}{5.2} & \meanwithstd{27.5}{3.2} & \meanwithstd{37.2}{1.4} & \meanwithstd{17.9}{4.2} & \meanwithstd{12.8}{1.8}\\
No irrelevant changes & \meanwithstd{22.8}{0.8} & \meanwithstd{67.1}{19.6} & \meanwithstd{33.7}{3.2} & \meanwithstd{4.2}{0.5} & \meanwithstd{54.2}{23.1} & \meanwithstd{7.8}{1.1} & \meanwithstd{6.2}{0.4} & \meanwithstd{75.3}{19.8} & \meanwithstd{11.5}{0.9} & \meanwithstd{6.5}{0.6} & \meanwithstd{63.8}{20.9} & \meanwithstd{11.7}{1.3} & \meanwithstd{7.3}{1.3} & \meanwithstd{92.5}{5.8} & \meanwithstd{13.5}{2.1} & \meanwithstd{9.4}{0.3} & \meanwithstd{70.6}{17.1} & \meanwithstd{15.6}{0.8}\\

    \bottomrule
    \end{tabular}
    }
\end{table*}

\begin{table*}[!h]
    \centering
        \caption{Different backbone ablation.}
    \label{atab:backbone}
    \resizebox{\linewidth}{!}{
\setlength{\tabcolsep}{2pt}
    \begin{tabular}{lccc|ccc|ccc|ccc|ccc|ccc}
    \toprule
    	\multirow{2}{*}{~}& \multicolumn{3}{c}{SYSU} & \multicolumn{3}{c}{LEVIR} & \multicolumn{3}{c}{GVLM} & \multicolumn{3}{c}{CLCD} & \multicolumn{3}{c}{OSCD} & \multicolumn{3}{c}{\textit{Avg.}} \\
& Pr. & Re. & F1& Pr. & Re. & F1& Pr. & Re. & F1& Pr. & Re. & F1& Pr. & Re. & F1& Pr. & Re. & F1\\
 \hline
ViT-L (DinoV3)\scriptsize{(Ours)} & \meanwithstd{72.3}{1.2} & \meanwithstd{52.3}{3.7} & \meanwithstd{60.6}{2.2} & \meanwithstd{25.9}{1.2} & \meanwithstd{78.5}{4.2} & \meanwithstd{38.9}{1.2} & \meanwithstd{43.2}{4.9} & \meanwithstd{78.5}{6.1} & \meanwithstd{55.4}{2.9} & \meanwithstd{46.3}{1.7} & \meanwithstd{66.4}{6.8} & \meanwithstd{54.3}{1.5} & \meanwithstd{34.2}{2.1} & \meanwithstd{60.3}{2.8} & \meanwithstd{43.5}{1.0} & \meanwithstd{44.4}{1.9} & \meanwithstd{67.2}{3.8} & \meanwithstd{50.6}{0.4}\\

ViT-L (DinoV2) & \meanwithstd{65.6}{3.5} & \meanwithstd{67.4}{9.3} & \meanwithstd{66.0}{3.0} & \meanwithstd{20.8}{1.5} & \meanwithstd{76.3}{11.0} & \meanwithstd{32.5}{0.8} & \meanwithstd{42.8}{6.1} & \meanwithstd{79.7}{9.2} & \meanwithstd{55.0}{3.3} & \meanwithstd{29.1}{3.9} & \meanwithstd{86.2}{4.8} & \meanwithstd{43.3}{3.7} & \meanwithstd{28.3}{4.0} & \meanwithstd{73.3}{7.4} & \meanwithstd{40.4}{2.9} & \meanwithstd{37.3}{3.6} & \meanwithstd{76.6}{8.0} & \meanwithstd{47.4}{1.5}\\
ViT-B (DinoV3) & \meanwithstd{73.7}{2.0} & \meanwithstd{43.2}{7.8} & \meanwithstd{54.0}{6.0} & \meanwithstd{25.4}{3.1} & \meanwithstd{84.3}{8.5} & \meanwithstd{38.7}{2.6} & \meanwithstd{37.9}{4.0} & \meanwithstd{45.2}{19.5} & \meanwithstd{38.7}{7.4} & \meanwithstd{43.9}{5.5} & \meanwithstd{64.1}{13.0} & \meanwithstd{51.1}{2.0} & \meanwithstd{40.6}{4.3} & \meanwithstd{50.5}{6.3} & \meanwithstd{44.6}{0.8} & \meanwithstd{44.3}{3.5} & \meanwithstd{57.5}{10.6} & \meanwithstd{45.4}{2.3}\\

Swin Tiny & \meanwithstd{51.9}{4.3} & \meanwithstd{65.1}{7.4} & \meanwithstd{57.4}{1.9} & \meanwithstd{16.2}{0.6} & \meanwithstd{94.9}{1.8} & \meanwithstd{27.7}{0.8} & \meanwithstd{26.3}{4.3} & \meanwithstd{73.3}{2.9} & \meanwithstd{38.5}{4.4} & \meanwithstd{24.2}{1.8} & \meanwithstd{67.2}{5.4} & \meanwithstd{35.5}{1.8} & \meanwithstd{38.7}{1.3} & \meanwithstd{67.8}{2.3} & \meanwithstd{49.3}{0.5} & \meanwithstd{31.5}{1.3} & \meanwithstd{73.7}{1.7} & \meanwithstd{41.7}{1.2}\\
ResNet50 & \meanwithstd{43.0}{1.0} & \meanwithstd{54.3}{3.3} & \meanwithstd{47.9}{0.8} & \meanwithstd{14.3}{0.8} & \meanwithstd{87.2}{5.0} & \meanwithstd{24.6}{1.1} & \meanwithstd{22.6}{0.8} & \meanwithstd{66.2}{5.6} & \meanwithstd{33.7}{1.0} & \meanwithstd{17.4}{0.5} & \meanwithstd{54.9}{6.8} & \meanwithstd{26.3}{1.0} & \meanwithstd{24.7}{2.0} & \meanwithstd{84.8}{2.8} & \meanwithstd{38.1}{2.0} & \meanwithstd{24.4}{0.6} & \meanwithstd{69.5}{3.2} & \meanwithstd{34.1}{0.5}\\
\bottomrule
    \end{tabular}
    }
\end{table*}

\begin{table*}[!h]
    \centering
        \caption{Different layer ablation.}
    \label{atab:layer}
    \resizebox{\linewidth}{!}{
\setlength{\tabcolsep}{2pt}
    \begin{tabular}{cccc ccc|ccc|ccc|ccc|ccc|ccc}
    \toprule
    	\multicolumn{4}{c}{Noise on layer}& \multicolumn{3}{c}{SYSU} & \multicolumn{3}{c}{LEVIR} & \multicolumn{3}{c}{GVLM} & \multicolumn{3}{c}{CLCD} & \multicolumn{3}{c}{OSCD} & \multicolumn{3}{c}{\textit{Avg.}} \\
7 & 11 & 15 & 23 & Pr. & Re. & F1& Pr. & Re. & F1& Pr. & Re. & F1& Pr. & Re. & F1& Pr. & Re. & F1& Pr. & Re. & F1\\
 \hline
$\checkmark$&$\checkmark$&$\checkmark$&$\checkmark$ & \meanwithstd{72.3}{1.2} & \meanwithstd{52.3}{3.7} & \meanwithstd{60.6}{2.2} & \meanwithstd{25.9}{1.2} & \meanwithstd{78.5}{4.2} & \meanwithstd{38.9}{1.2} & \meanwithstd{43.2}{4.9} & \meanwithstd{78.5}{6.1} & \meanwithstd{55.4}{2.9} & \meanwithstd{46.3}{1.7} & \meanwithstd{66.4}{6.8} & \meanwithstd{54.3}{1.5} & \meanwithstd{34.2}{2.1} & \meanwithstd{60.3}{2.8} & \meanwithstd{43.5}{1.0} & \meanwithstd{44.4}{1.9} & \meanwithstd{67.2}{3.8} & \meanwithstd{50.6}{0.4}\\

$\checkmark$&$\checkmark$&$\checkmark$&~ & \meanwithstd{70.5}{1.6} & \meanwithstd{53.0}{3.8} & \meanwithstd{60.4}{1.9} & \meanwithstd{26.4}{1.2} & \meanwithstd{72.0}{8.6} & \meanwithstd{38.5}{1.3} & \meanwithstd{36.5}{2.6} & \meanwithstd{84.5}{4.4} & \meanwithstd{50.9}{2.0} & \meanwithstd{43.2}{1.7} & \meanwithstd{70.6}{1.6} & \meanwithstd{53.6}{1.1} & \meanwithstd{27.8}{2.1} & \meanwithstd{71.7}{3.8} & \meanwithstd{40.0}{1.6} & \meanwithstd{40.9}{1.1} & \meanwithstd{70.3}{3.2} & \meanwithstd{48.6}{0.3}\\
$\checkmark$&$\checkmark$&~&$\checkmark$ & \meanwithstd{63.9}{2.3} & \meanwithstd{71.9}{3.9} & \meanwithstd{67.5}{0.5} & \meanwithstd{20.3}{1.9} & \meanwithstd{94.3}{2.1} & \meanwithstd{33.3}{2.5} & \meanwithstd{23.4}{2.3} & \meanwithstd{94.5}{1.7} & \meanwithstd{37.4}{2.8} & \meanwithstd{33.1}{1.5} & \meanwithstd{85.9}{1.4} & \meanwithstd{47.8}{1.4} & \meanwithstd{24.5}{2.3} & \meanwithstd{77.2}{3.8} & \meanwithstd{37.1}{2.4} & \meanwithstd{33.0}{1.9} & \meanwithstd{84.8}{2.4} & \meanwithstd{44.6}{1.6}\\
$\checkmark$&~&$\checkmark$&$\checkmark$ & \meanwithstd{66.5}{1.6} & \meanwithstd{67.4}{3.4} & \meanwithstd{66.9}{1.0} & \meanwithstd{21.4}{2.0} & \meanwithstd{92.1}{3.2} & \meanwithstd{34.7}{2.4} & \meanwithstd{27.8}{2.9} & \meanwithstd{92.2}{2.5} & \meanwithstd{42.6}{3.3} & \meanwithstd{34.9}{1.7} & \meanwithstd{84.1}{1.9} & \meanwithstd{49.3}{1.4} & \meanwithstd{25.2}{2.4} & \meanwithstd{75.0}{4.1} & \meanwithstd{37.6}{2.2} & \meanwithstd{35.2}{1.9} & \meanwithstd{82.1}{2.8} & \meanwithstd{46.2}{1.6}\\
~&$\checkmark$&$\checkmark$&$\checkmark$ & \meanwithstd{69.8}{1.5} & \meanwithstd{60.1}{4.3} & \meanwithstd{64.5}{1.9} & \meanwithstd{23.4}{1.9} & \meanwithstd{87.1}{5.3} & \meanwithstd{36.8}{1.9} & \meanwithstd{34.7}{4.0} & \meanwithstd{85.8}{5.6} & \meanwithstd{49.2}{3.2} & \meanwithstd{40.0}{1.5} & \meanwithstd{78.0}{2.9} & \meanwithstd{52.9}{0.8} & \meanwithstd{28.9}{2.9} & \meanwithstd{67.8}{5.1} & \meanwithstd{40.3}{2.1} & \meanwithstd{39.4}{2.2} & \meanwithstd{75.8}{4.2} & \meanwithstd{48.7}{1.2}\\

    \bottomrule
    \end{tabular}
    }
\end{table*}

\begin{table*}[!h]
    \centering
        \caption{Noise sampling dimension ablation.}
    \label{atab:qdim}
    \resizebox{\linewidth}{!}{
\setlength{\tabcolsep}{2pt}
    \begin{tabular}{lccc|ccc|ccc|ccc|ccc|ccc}
    \toprule
    	\multirow{2}{*}{~}& \multicolumn{3}{c}{SYSU} & \multicolumn{3}{c}{LEVIR} & \multicolumn{3}{c}{GVLM} & \multicolumn{3}{c}{CLCD} & \multicolumn{3}{c}{OSCD} & \multicolumn{3}{c}{\textit{Avg.}} \\
& Pr. & Re. & F1& Pr. & Re. & F1& Pr. & Re. & F1& Pr. & Re. & F1& Pr. & Re. & F1& Pr. & Re. & F1\\
 \hline
Per channel in batch & \meanwithstd{72.3}{1.2} & \meanwithstd{52.3}{3.7} & \meanwithstd{60.6}{2.2} & \meanwithstd{25.9}{1.2} & \meanwithstd{78.5}{4.2} & \meanwithstd{38.9}{1.2} & \meanwithstd{43.2}{4.9} & \meanwithstd{78.5}{6.1} & \meanwithstd{55.4}{2.9} & \meanwithstd{46.3}{1.7} & \meanwithstd{66.4}{6.8} & \meanwithstd{54.3}{1.5} & \meanwithstd{34.2}{2.1} & \meanwithstd{60.3}{2.8} & \meanwithstd{43.5}{1.0} & \meanwithstd{44.4}{1.9} & \meanwithstd{67.2}{3.8} & \meanwithstd{50.6}{0.4}\\

Per channel in sample & \meanwithstd{74.8}{2.9} & \meanwithstd{14.9}{8.1} & \meanwithstd{24.2}{10.7} & \meanwithstd{28.3}{4.6} & \meanwithstd{14.0}{17.3} & \meanwithstd{15.3}{13.1} & \meanwithstd{59.0}{3.9} & \meanwithstd{27.3}{18.0} & \meanwithstd{34.8}{16.0} & \meanwithstd{53.4}{2.3} & \meanwithstd{25.7}{12.3} & \meanwithstd{33.4}{10.9} & \meanwithstd{50.2}{6.3} & \meanwithstd{35.4}{10.5} & \meanwithstd{40.2}{4.2} & \meanwithstd{53.1}{1.6} & \meanwithstd{23.5}{13.0} & \meanwithstd{29.6}{10.7}\\
Per sample & \meanwithstd{64.0}{2.6} & \meanwithstd{49.8}{10.6} & \meanwithstd{55.4}{5.5} & \meanwithstd{23.0}{2.3} & \meanwithstd{67.2}{16.8} & \meanwithstd{33.9}{2.8} & \meanwithstd{29.5}{6.1} & \meanwithstd{75.9}{13.6} & \meanwithstd{41.4}{4.8} & \meanwithstd{36.9}{2.6} & \meanwithstd{68.2}{10.2} & \meanwithstd{47.5}{1.6} & \meanwithstd{37.1}{6.4} & \meanwithstd{55.6}{10.5} & \meanwithstd{43.4}{2.0} & \meanwithstd{38.1}{3.8} & \meanwithstd{63.4}{11.7} & \meanwithstd{44.3}{0.8}\\
Per batch & \meanwithstd{45.7}{2.4} & \meanwithstd{84.3}{3.1} & \meanwithstd{59.2}{1.4} & \meanwithstd{13.8}{1.3} & \meanwithstd{97.2}{1.1} & \meanwithstd{24.1}{2.0} & \meanwithstd{9.7}{0.8} & \meanwithstd{98.2}{1.0} & \meanwithstd{17.7}{1.4} & \meanwithstd{17.7}{0.9} & \meanwithstd{93.3}{1.2} & \meanwithstd{29.7}{1.3} & \meanwithstd{21.6}{2.7} & \meanwithstd{82.4}{4.4} & \meanwithstd{34.0}{3.1} & \meanwithstd{21.7}{1.6} & \meanwithstd{91.1}{2.1} & \meanwithstd{32.9}{1.7}\\

    \bottomrule
    \end{tabular}
    }
\end{table*}

\section{Additional Qualitative Results}
\label{a:qual}

In this section, we provide additional qualitative examples of MaSoN and the related methods: pixel-differencing, DINOv3 CVA~\cite{simeoni2025dinov3, zheng2024anychange}, I3PE~\cite{chen2023_i3pe}, HySCDG~\cite{benidir2025hyscdg}, Changen2~\cite{zgeng2025changen2}, AnyChange~\cite{zheng2024anychange}, DynamicEarth~\cite{li2025dynearth}, and S2C~\cite{ding2025s2c}. Compared to related work, MaSoN more accurately predicts actual changes and is more robust towards irrelevant changes.

\begin{figure*}
    \centering
    \includegraphics[width=1\linewidth]{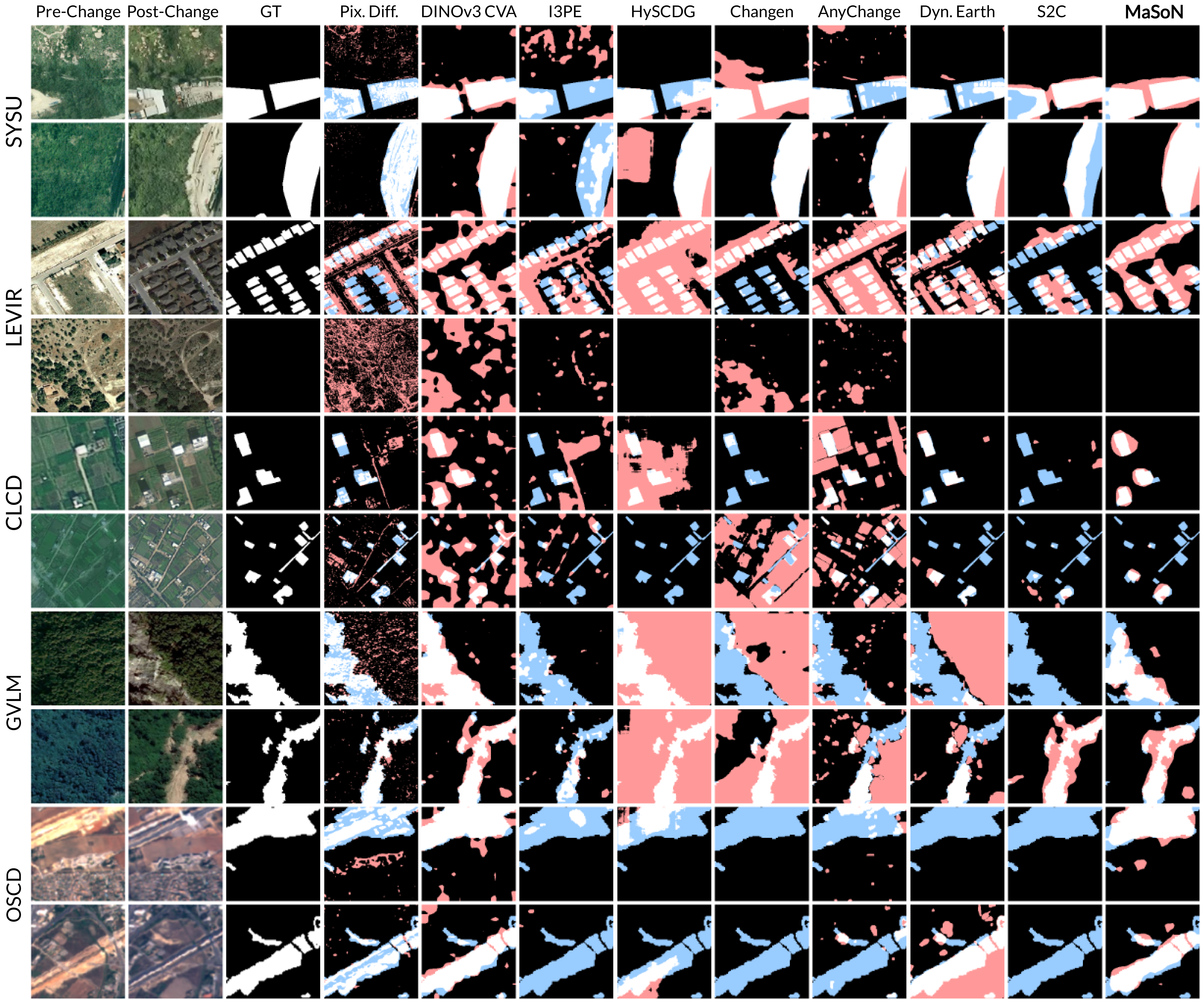}
    \caption{Qualitative comparison of the predictions made by our method and related methods. The pair of considered images is shown in the first and second columns, followed by the ground truth mask and predictions for each method. \textcolor{falsePos}{False positives are marked in red} and \textcolor{falseNeg}{false negatives in blue}.}
    \label{fig:extra_qual}
\end{figure*}

\begin{figure*}
    \centering
    \includegraphics[width=1\linewidth]{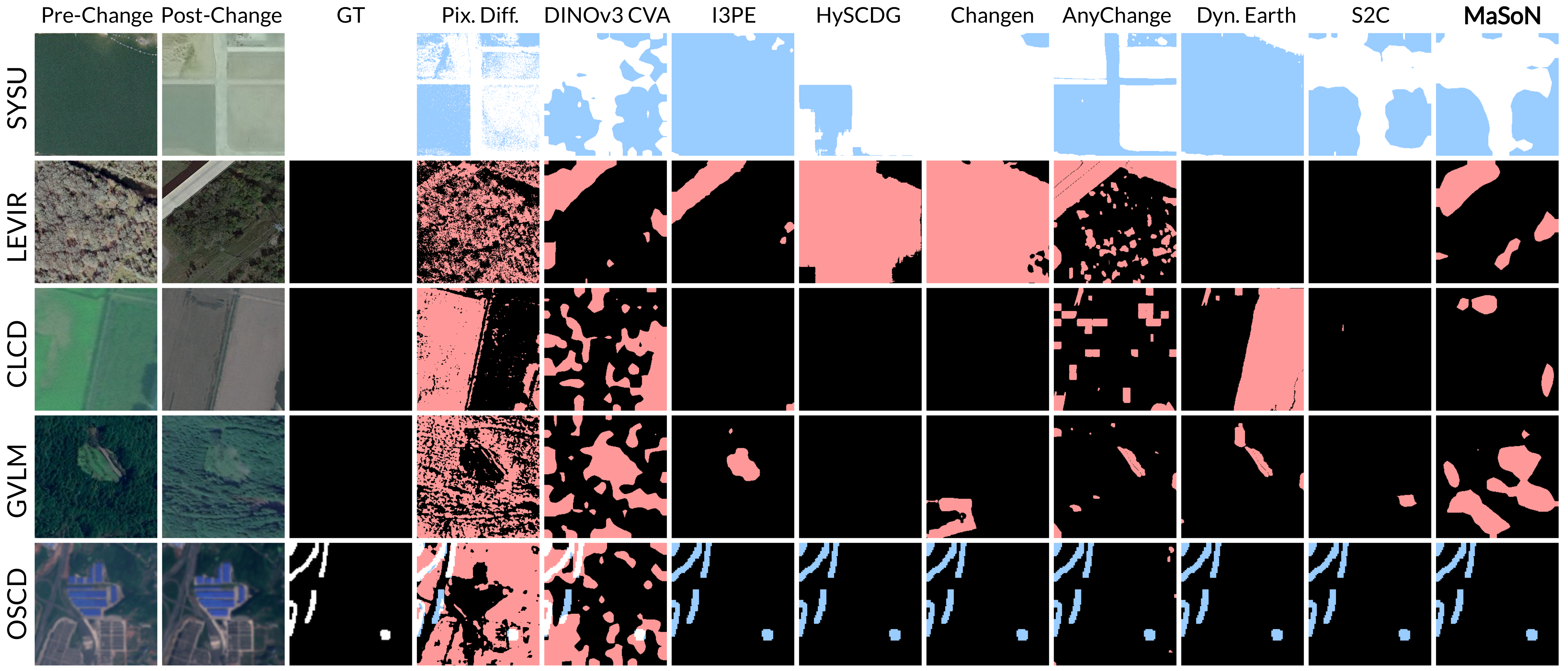}
    \caption{Failure cases made by our method and related methods. }
    \label{fig:failure}
\end{figure*}

\subsection{Failure Cases}

In this section, we highlight a few typical failure cases. MaSoN struggles in ambiguous scenarios where the definition of "change" is context-dependent. For instance, in the third row (CLCD), it predicts a change due to visible vegetation growth in a field. While this could reasonably be considered a change, it is not considered in this particular dataset. A similar observation can be made in the case of LEVIR (second row), where only buildings are labelled, but many unsupervised methods detect all changes, therefore marking the construction of the road as a change. Such cases reflect the inherent limitations of unsupervised learning. However, these ambiguities could be mitigated by filtering (results for this were presented in the main paper \Cref{sec:abl}) or finetuning the model with a small number of labelled examples.

\section{Supervised Results}
\label{a:sup}

MaSoN can also be trained in a supervised fashion. The results for this setting are presented in~\Cref{atab:sup}. Interestingly, MaSoN achieves only 10 p.p. higher recall to the one achieved in the unsupervised setting. Contrary to the unsupervised setting, the model achieves a much higher Precision. This likely happens due to the nature of most datasets, where only certain changes, like buildings, need to be detected, a task that requires at least some sort of supervision (or filtering) to convey what a building is. The other reason is that since we generate changes in the unsupervised case at a smaller resolution (i.e., that of feature maps, which are 16 times smaller than image for ViT), supervision targets don't offer the same precision, leading to more blobby predicted masks.

\begin{table*}[!ht]
    \centering
    \caption{Supervised change detection results across five diverse datasets and their average. We report Precision (Pr.), Recall (Re.), and F1 score.}
    \label{atab:sup}
    \resizebox{\linewidth}{!}{
\setlength{\tabcolsep}{2pt}
    \begin{tabular}{lccc|ccc|ccc|ccc|ccc|ccc}
    \toprule
    	\multirow{2}{*}{~}& \multicolumn{3}{c}{SYSU} & \multicolumn{3}{c}{LEVIR} & \multicolumn{3}{c}{GVLM} & \multicolumn{3}{c}{CLCD} & \multicolumn{3}{c}{OSCD} & \multicolumn{3}{c}{\textit{Avg}} \\
& Pr. & Re. & F1& Pr. & Re. & F1& Pr. & Re. & F1& Pr. & Re. & F1& Pr. & Re. & F1& Pr. & Re. & F1\\
 \hline
FCS-Diff~\cite{daudt2018fcn} & \meanwithstd{83.5}{0.4} & \meanwithstd{61.5}{0.3} & \meanwithstd{70.8}{0.1} & \meanwithstd{83.0}{0.9} & \meanwithstd{80.6}{2.0} & \meanwithstd{81.8}{0.7} & \meanwithstd{82.7}{3.1} & \meanwithstd{67.6}{2.5} & \meanwithstd{74.3}{0.8} & \meanwithstd{54.0}{1.0} & \meanwithstd{54.3}{2.6} & \meanwithstd{54.1}{1.2} & \meanwithstd{27.8}{1.1} & \meanwithstd{\bm1{68.5}}{6.8} & \meanwithstd{39.4}{1.4} & \meanwithstd{66.2}{1.0} & \meanwithstd{66.5}{2.2} & \meanwithstd{64.1}{0.4}\\
BIT~\cite{chen2021bit} & \meanwithstd{79.1}{2.3} & \meanwithstd{75.1}{2.1} & \meanwithstd{77.0}{0.1} & \meanwithstd{92.0}{0.4} & \meanwithstd{88.1}{0.2} & \meanwithstd{90.0}{0.1} & \meanwithstd{89.8}{1.2} & \meanwithstd{87.9}{1.1} & \meanwithstd{88.8}{0.1} & \meanwithstd{66.5}{5.8} & \meanwithstd{60.8}{1.3} & \meanwithstd{63.4}{2.0} & \meanwithstd{48.7}{1.9} & \meanwithstd{37.9}{2.2} & \meanwithstd{42.6}{2.0} & \meanwithstd{75.2}{2.2} & \meanwithstd{70.0}{0.8} & \meanwithstd{72.4}{0.6}\\
ChFormer~\cite{bandara2022changeFormer} & \meanwithstd{82.8}{0.4} & \meanwithstd{73.5}{1.3} & \meanwithstd{77.9}{0.5} & \meanwithstd{91.7}{0.2} & \meanwithstd{87.3}{0.5} & \meanwithstd{89.5}{0.2} & \meanwithstd{90.4}{0.6} & \meanwithstd{86.6}{0.4} & \meanwithstd{88.5}{0.2} & \meanwithstd{61.4}{1.3} & \meanwithstd{60.4}{5.1} & \meanwithstd{60.8}{2.6} & \meanwithstd{60.2}{0.5} & \meanwithstd{40.1}{1.1} & \meanwithstd{48.1}{1.0} & \meanwithstd{77.3}{0.2} & \meanwithstd{69.6}{1.1} & \meanwithstd{73.0}{0.5}\\
BiFA~\cite{zhang2024bifa} & \meanwithstd{87.4}{0.2} & \meanwithstd{\bm1{80.4}}{0.6} & \meanwithstd{\bm1{83.8}}{0.3} & \meanwithstd{90.9}{1.1} & \meanwithstd{88.1}{0.9} & \meanwithstd{89.5}{0.9} & \meanwithstd{89.9}{0.7} & \meanwithstd{88.2}{1.0} & \meanwithstd{89.0}{0.2} & \meanwithstd{79.4}{2.3} & \meanwithstd{70.1}{0.7} & \meanwithstd{74.5}{0.6} & \meanwithstd{61.5}{2.3} & \meanwithstd{27.1}{4.4} & \meanwithstd{37.4}{3.7} & \meanwithstd{81.8}{0.7} & \meanwithstd{70.8}{1.1} & \meanwithstd{74.8}{0.9}\\
SwinSUNet~\cite{zhang2022swinsunet} & \meanwithstd{89.2}{0.6} & \meanwithstd{67.2}{2.3} & \meanwithstd{76.6}{1.7} & \meanwithstd{86.9}{0.1} & \meanwithstd{\bm1{91.7}}{0.3} & \meanwithstd{89.3}{0.1} & \meanwithstd{88.2}{1.8} & \meanwithstd{\bm1{92.0}}{1.0} & \meanwithstd{90.0}{0.5} & \meanwithstd{79.5}{1.6} & \meanwithstd{72.5}{2.6} & \meanwithstd{75.8}{0.8} & \meanwithstd{\bm2{61.7}}{1.3} & \meanwithstd{46.3}{5.3} & \meanwithstd{52.8}{3.0} & \meanwithstd{81.1}{0.9} & \meanwithstd{73.9}{1.4} & \meanwithstd{76.9}{0.4}\\
ChangeMamba~\cite{chen2024changeMamba} & \meanwithstd{89.6}{0.3} & \meanwithstd{74.7}{0.6} & \meanwithstd{81.5}{0.3} & \meanwithstd{92.4}{0.2} & \meanwithstd{\bm2{91.2}}{0.1} & \meanwithstd{\bm2{91.8}}{0.1} & \meanwithstd{\bm2{91.2}}{0.9} & \meanwithstd{\bm2{90.4}}{0.8} & \meanwithstd{\bm1{90.8}}{0.1} & \meanwithstd{\bm1{87.3}}{0.6} & \meanwithstd{74.4}{1.5} & \meanwithstd{\bm2{80.3}}{1.1} & \meanwithstd{\bm1{63.4}}{3.7} & \meanwithstd{36.1}{3.2} & \meanwithstd{45.8}{1.8} & \meanwithstd{\bm1{84.8}}{0.5} & \meanwithstd{73.4}{0.6} & \meanwithstd{78.0}{0.5}\\
CaCo~\cite{mall2023caco} & \meanwithstd{88.4}{0.7} & \meanwithstd{73.4}{0.3} & \meanwithstd{80.2}{0.4} & \meanwithstd{92.0}{0.1} & \meanwithstd{89.9}{0.2} & \meanwithstd{90.9}{0.0} & \meanwithstd{\bm1{91.4}}{0.1} & \meanwithstd{89.2}{0.2} & \meanwithstd{\bm2{90.3}}{0.0} & \meanwithstd{83.9}{0.8} & \meanwithstd{72.6}{1.2} & \meanwithstd{77.8}{0.5} & \meanwithstd{54.5}{4.1} & \meanwithstd{49.8}{2.8} & \meanwithstd{51.9}{0.9} & \meanwithstd{82.0}{0.6} & \meanwithstd{75.0}{0.4} & \meanwithstd{78.2}{0.0}\\
SatMAE~\cite{cong2022satmae} & \meanwithstd{89.0}{0.0} & \meanwithstd{75.0}{0.1} & \meanwithstd{81.4}{0.1} & \meanwithstd{92.6}{0.1} & \meanwithstd{90.2}{0.1} & \meanwithstd{91.4}{0.1} & \meanwithstd{90.2}{0.2} & \meanwithstd{89.3}{0.4} & \meanwithstd{89.8}{0.2} & \meanwithstd{84.0}{1.2} & \meanwithstd{74.9}{0.5} & \meanwithstd{79.2}{0.8} & \meanwithstd{53.8}{1.5} & \meanwithstd{45.9}{2.7} & \meanwithstd{49.5}{1.2} & \meanwithstd{82.0}{0.3} & \meanwithstd{75.1}{0.6} & \meanwithstd{78.3}{0.2}\\
GFM~\cite{mendieta2023gfm} & \meanwithstd{\bm2{89.7}}{0.3} & \meanwithstd{74.3}{0.6} & \meanwithstd{81.2}{0.2} & \meanwithstd{90.8}{0.2} & \meanwithstd{88.8}{0.1} & \meanwithstd{89.8}{0.1} & \meanwithstd{90.5}{0.1} & \meanwithstd{89.2}{0.1} & \meanwithstd{89.8}{0.1} & \meanwithstd{82.2}{1.0} & \meanwithstd{73.2}{0.9} & \meanwithstd{77.5}{0.9} & \meanwithstd{55.9}{0.5} & \meanwithstd{52.5}{2.1} & \meanwithstd{\bm2{54.1}}{1.4} & \meanwithstd{81.8}{0.3} & \meanwithstd{75.6}{0.5} & \meanwithstd{78.5}{0.4}\\
MTP~\cite{wang2024mtp} & \meanwithstd{88.5}{0.5} & \meanwithstd{75.2}{0.9} & \meanwithstd{81.3}{0.3} & \meanwithstd{\bm2{92.8}}{0.1} & \meanwithstd{90.7}{0.1} & \meanwithstd{91.7}{0.0} & \meanwithstd{90.8}{0.5} & \meanwithstd{89.0}{0.4} & \meanwithstd{89.9}{0.2} & \meanwithstd{85.4}{0.9} & \meanwithstd{\bm2{75.8}}{0.3} & \meanwithstd{\bm2{80.3}}{0.3} & \meanwithstd{43.9}{2.4} & \meanwithstd{\bm2{66.5}}{4.4} & \meanwithstd{52.8}{0.4} & \meanwithstd{80.3}{0.7} & \meanwithstd{\bm1{79.4}}{1.0} & \meanwithstd{\bm2{79.2}}{0.1}\\
\textbf{Ours} & \meanwithstd{\bm1{90.7}}{0.1} & \meanwithstd{\bm2{77.6}}{0.6} & \meanwithstd{\bm2{83.7}}{0.4} & \meanwithstd{\bm1{92.9}}{0.1} & \meanwithstd{\bm2{91.2}}{0.1} & \meanwithstd{\bm1{92.1}}{0.1} & \meanwithstd{91.1}{0.2} & \meanwithstd{89.4}{0.2} & \meanwithstd{\bm2{90.3}}{0.2} & \meanwithstd{\bm2{86.8}}{0.9} & \meanwithstd{\bm1{77.3}}{1.0} & \meanwithstd{\bm1{81.8}}{0.7} & \meanwithstd{57.0}{0.8} & \meanwithstd{57.5}{3.9} & \meanwithstd{\bm1{57.2}}{1.9} & \meanwithstd{\bm2{83.7}}{0.3} & \meanwithstd{\bm2{78.6}}{0.6} & \meanwithstd{\bm1{81.0}}{0.2}\\

    \bottomrule
    \end{tabular}
    }
\end{table*}

\section{Computational Efficiency}
\label{a:comp}

Implementation of our protocol for measuring computational efficiency is described here (\Cref{asub:per_prot}). Some additional metrics, such as GFLOPs and inference time, are also provided for our method and other state-of-the-art methods (\Cref{asub:addit_comp_res}). Finally, we also discuss the computational overhead of noise generation in \Cref{asub:noise_overhead}.

\subsection{Protocol Implementation Details}
\label{asub:per_prot}

3 different computational efficiency metrics are reported: parameter count, inference time (also expressed as frames per second - FPS) and GFLOPs. GFLOPs are measured using the official PyTorch profiler. 

For inference time, we use a pair of $256 \times 256$ RGB float16 images as input. Models are cast to float16 where supported. We perform 1000 warm-up forward passes, followed by 1000 timed forward passes. This procedure is repeated five times, and the average runtime per forward pass is reported. All measurements are conducted on an NVIDIA A100-SXM4 40GB GPU.

\subsection{Additional Results}
\label{asub:addit_comp_res}

The results for unsupervised methods are reported in Table~\ref{atab:unsup_perf}. Our method outperforms the second-best DynamicEarth~\cite{li2025dynearth} by 14.4 percentage points of F1 score, while having half the parameters and offering significantly faster inference speed. Pixel differencing still runs significantly faster than any other method and has the benefit of having no parameters. However, the performance it achieves is substantially below the best deep learning methods.

We can also see that DINO-based CVA is slower than our model. This comes from the fact that CVA is a post-processing step on DINO features, essentially replacing our UPerNet decoder. It relies on norm calculation and Otsu threshold, which are not as fast as a simple forward decoder pass. This shows the benefit of fully end-to-end solutions.

\begin{table}[!h]
    \centering
        \caption{Computational efficiency results for each model. We report FPS (derived from inference time), inference time, parameter count, GFLOPs, and average Precision, Recall, and F1 across 5 datasets. All results were obtained using an Nvidia A100-SXM4 40GB GPU.}
        \label{atab:unsup_perf}
        \resizebox{\linewidth}{!}{
    \setlength{\tabcolsep}{2pt}
    \begin{tabular}{lccccccc}
    \toprule
    	\multirow{2}{*}{~} & FPS & Inference Time & Param. & FLOPS& \multicolumn{3}{c}{\textit{Avg. Change Detection }} \\
~ & [img/s] & [ms] & [M] & [$10^9$]& Pr. & Re. & F1\\
 \hline
Pixel Difference & 456.4{\scriptsize$\pm54.5$} & 2.2{\scriptsize$\pm0.2$} & 0.0 & 0.0 & 16.2 & 49.3 & 22.0\\
DCVA~\cite{saha2019unsupervised}\tiny{TGRS19} & 0.4{\scriptsize$\pm0.0$} & 2824.9{\scriptsize$\pm6.1$} & 0.5 & 21.9 & 30.0 & 9.3 & 9.6\\
DINOv2-CVA & 24.4{\scriptsize$\pm0.3$} & 41.0{\scriptsize$\pm0.5$} & 303.1 & 316.2 & 15.7 & \bm1{ 85.1} & 25.0\\
DINOv3-CVA & 24.4{\scriptsize$\pm0.3$} & 41.0{\scriptsize$\pm0.5$} & 303.1 & 316.2 & 19.0 & \bm2{ 80.4} & 28.7\\
AnyChange~\cite{zheng2024anychange}\tiny{NeurIPS24} & 0.4{\scriptsize$\pm0.0$} & 2234.7{\scriptsize$\pm0.2$} & 641.1 & 5789.3 & 26.0 & 43.0 & 29.0\\
Changen2-S9~\cite{zgeng2025changen2}\tiny{TPAMI25} & 33.3{\scriptsize$\pm0.1$} & 30.0{\scriptsize$\pm0.1$} & 99.9 & 197.6 & 17.3 & 39.4 & 20.2\\
CDRL*~\cite{noh2022cdrl}\tiny{CVPRW22} & 130.8{\scriptsize$\pm0.1$} & 7.6{\scriptsize$\pm0.0$} & 54.5 & 8.9 & 11.2{\scriptsize$\pm0.2$} & 3.1{\scriptsize$\pm0.0$} & 3.7{\scriptsize$\pm0.0$}\\
CDRL-SA*~\cite{noh2024cdrlSA}\tiny{RSL24} & 0.3{\scriptsize$\pm0.0$} & 3055.1{\scriptsize$\pm0.8$} & 682.5 & 17742.1 & 28.4{\scriptsize$\pm4.2$} & 1.0{\scriptsize$\pm0.2$} & 1.9{\scriptsize$\pm0.3$}\\
I3PE~\cite{chen2023_i3pe}\tiny{ISPRS23} & 65.9{\scriptsize$\pm0.2$} & 15.2{\scriptsize$\pm0.0$} & 24.9 & 35.2 & 15.6{\scriptsize$\pm0.3$} & 24.8{\scriptsize$\pm3.1$} & 17.2{\scriptsize$\pm1.6$}\\
HySCDG~\cite{benidir2025hyscdg}\tiny{CVPR25} & 41.0{\scriptsize$\pm0.1$} & 24.4{\scriptsize$\pm0.0$} & 65.1 & 64.8 & 29.2 & 65.2 & 25.5\\
SCM~\cite{tan2024scm}\tiny{IGARSS24} & 1.2{\scriptsize$\pm0.0$} & 817.0{\scriptsize$\pm16.8$} & 223.5 & 423.2 & 21.6 & 30.1 & 23.3\\
DynEarth~\cite{li2025dynearth}\tiny{AAAI26} & 0.3{\scriptsize$\pm0.0$} & 3676.7{\scriptsize$\pm0.5$} & 877.6 & 19541.3 & 30.8 & 45.9 & 35.0\\
DynE. (DINOv3)~\cite{li2025dynearth}\tiny{AAAI26} & 0.3{\scriptsize$\pm0.0$} & 3702.8{\scriptsize$\pm0.8$} & 877.6 & 19542.6 & 34.9 & 42.0 & 36.2\\
S2C (DINOv3)~\cite{ding2025s2c}\tiny{AAAI26} & 3.8{\scriptsize$\pm0.0$} & 260.7{\scriptsize$\pm0.6$} & 303.6 & 1266.8 & \bm2{41.4} {\scriptsize$\pm4.2$} & 42.1{\scriptsize$\pm5.9$} & \bm2{36.5} {\scriptsize$\pm4.5$}\\
\textbf{Ours} & 39.7{\scriptsize$\pm0.1$} & 25.2{\scriptsize$\pm0.1$} & 337.5 & 332.8 & \meanwithstd{\bm1{44.4}}{1.9} & \meanwithstd{67.2}{3.8} & \meanwithstd{\bm1{50.6}}{0.4}\\

    \bottomrule
    \end{tabular}
    }
\end{table}

\subsection{Discussion on Noise Generation Overhead}
\label{asub:noise_overhead}

The noise generation strategy introduces a small training overhead of approximately 20 milliseconds per single pass. As mentioned in the main paper, the entire training run for each dataset lasts only 7 minutes. This is relatively little compared to the hundreds of hours needed to train diffusion models like HySCDG~\cite{benidir2025hyscdg} (as per Appendix A of HySCDG paper).

\section{Additional Ablation Studies}
\label{a:add_abl}

\paragraph{Effect of Batch Size} The calculation of irrelevant and relevant noise relies on the images within a batch. This outcome is dependent on the batch size; therefore, we examine how batch size influences downstream performance. The findings are presented in Table~\ref{atab:bs}. It is evident that MaSoN demonstrates a reasonable robustness to this parameter.

\begin{table*}[!ht]
    \centering
\caption{Effect of batch size ablation.}
    \label{atab:bs}
    \resizebox{\linewidth}{!}{
\setlength{\tabcolsep}{2pt}
    \begin{tabular}{lccc|ccc|ccc|ccc|ccc|ccc}
    \toprule
    	\multirow{2}{*}{~}& \multicolumn{3}{c}{SYSU} & \multicolumn{3}{c}{LEVIR} & \multicolumn{3}{c}{GVLM} & \multicolumn{3}{c}{CLCD} & \multicolumn{3}{c}{OSCD} & \multicolumn{3}{c}{\textit{Avg.}} \\
& Pr. & Re. & F1& Pr. & Re. & F1& Pr. & Re. & F1& Pr. & Re. & F1& Pr. & Re. & F1& Pr. & Re. & F1\\
 \hline
BS 16 \scriptsize{(Ours)} & \meanwithstd{72.3}{1.2} & \meanwithstd{52.3}{3.7} & \meanwithstd{60.6}{2.2} & \meanwithstd{25.9}{1.2} & \meanwithstd{78.5}{4.2} & \meanwithstd{38.9}{1.2} & \meanwithstd{43.2}{4.9} & \meanwithstd{78.5}{6.1} & \meanwithstd{55.4}{2.9} & \meanwithstd{46.3}{1.7} & \meanwithstd{66.4}{6.8} & \meanwithstd{54.3}{1.5} & \meanwithstd{34.2}{2.1} & \meanwithstd{60.3}{2.8} & \meanwithstd{43.5}{1.0} & \meanwithstd{44.4}{1.9} & \meanwithstd{67.2}{3.8} & \meanwithstd{50.6}{0.4}\\

BS 8 & \meanwithstd{72.2}{3.4} & \meanwithstd{54.7}{7.7} & \meanwithstd{61.8}{3.3} & \meanwithstd{25.3}{1.5} & \meanwithstd{80.4}{5.9} & \meanwithstd{38.4}{1.3} & \meanwithstd{39.1}{6.1} & \meanwithstd{82.2}{7.4} & \meanwithstd{52.4}{4.5} & \meanwithstd{37.5}{5.2} & \meanwithstd{78.9}{7.0} & \meanwithstd{50.4}{3.5} & \meanwithstd{31.8}{4.0} & \meanwithstd{63.1}{6.8} & \meanwithstd{41.9}{2.3} & \meanwithstd{41.2}{3.8} & \meanwithstd{71.9}{6.2} & \meanwithstd{49.0}{1.6}\\
BS 32 & \meanwithstd{74.5}{1.7} & \meanwithstd{45.3}{7.0} & \meanwithstd{56.0}{4.7} & \meanwithstd{27.6}{1.3} & \meanwithstd{69.6}{9.9} & \meanwithstd{39.3}{1.1} & \meanwithstd{46.2}{6.2} & \meanwithstd{72.1}{10.9} & \meanwithstd{55.5}{2.1} & \meanwithstd{46.2}{2.4} & \meanwithstd{64.2}{8.2} & \meanwithstd{53.4}{1.6} & \meanwithstd{36.0}{3.1} & \meanwithstd{58.3}{4.7} & \meanwithstd{44.3}{0.9} & \meanwithstd{46.1}{2.8} & \meanwithstd{61.9}{7.9} & \meanwithstd{49.7}{0.9}\\

    \bottomrule
    \end{tabular}
    }
\end{table*}

\subsection{Relevant noise sampling source}

Our model samples relevant noise from the concatenation of features, but it can alternatively sample from the feature difference, as is the case for irrelevant noise. We evaluate this variant in \Cref{atab:source}. While sampling from the feature difference is necessary to capture intra-image variability, we find that for relevant noise, it is more effective to sample directly from the features. We hypothesise that this is because relevant changes should manifest as higher-magnitude variations, yet remain constrained within the realistic bounds of the underlying feature distribution. However, the difference in results is only minor, indicating that both options are viable as long as the sampled distribution is dynamically scaled and broader than the distribution of irrelevant changes (refer to \Cref{sub:feat_analysis}).

\begin{table*}[!ht]
    \centering
        \caption{Relevant noise statistics sampling source results.}
    \label{atab:source}
    \resizebox{\linewidth}{!}{
\setlength{\tabcolsep}{2pt}
    \begin{tabular}{lccc|ccc|ccc|ccc|ccc|ccc}
    \toprule
    	\multirow{2}{*}{~}& \multicolumn{3}{c}{SYSU} & \multicolumn{3}{c}{LEVIR} & \multicolumn{3}{c}{GVLM} & \multicolumn{3}{c}{CLCD} & \multicolumn{3}{c}{OSCD} & \multicolumn{3}{c}{\textit{Avg.}} \\
& Pr. & Re. & F1& Pr. & Re. & F1& Pr. & Re. & F1& Pr. & Re. & F1& Pr. & Re. & F1& Pr. & Re. & F1\\
 \hline
 Feat. concat\scriptsize{(Ours)} & \meanwithstd{72.3}{1.2} & \meanwithstd{52.3}{3.7} & \meanwithstd{60.6}{2.2} & \meanwithstd{25.9}{1.2} & \meanwithstd{78.5}{4.2} & \meanwithstd{38.9}{1.2} & \meanwithstd{43.2}{4.9} & \meanwithstd{78.5}{6.1} & \meanwithstd{55.4}{2.9} & \meanwithstd{46.3}{1.7} & \meanwithstd{66.4}{6.8} & \meanwithstd{54.3}{1.5} & \meanwithstd{34.2}{2.1} & \meanwithstd{60.3}{2.8} & \meanwithstd{43.5}{1.0} & \meanwithstd{44.4}{1.9} & \meanwithstd{67.2}{3.8} & \meanwithstd{50.6}{0.4}\\

Feat. diff. & \meanwithstd{75.2}{1.5} & \meanwithstd{45.9}{6.6} & \meanwithstd{56.7}{4.8} & \meanwithstd{27.5}{2.3} & \meanwithstd{66.9}{12.5} & \meanwithstd{38.6}{1.7} & \meanwithstd{48.3}{6.1} & \meanwithstd{69.7}{12.0} & \meanwithstd{56.1}{0.8} & \meanwithstd{45.6}{2.9} & \meanwithstd{66.8}{8.2} & \meanwithstd{53.9}{1.4} & \meanwithstd{26.6}{3.1} & \meanwithstd{72.9}{5.2} & \meanwithstd{38.8}{2.8} & \meanwithstd{44.6}{2.8} & \meanwithstd{64.5}{8.4} & \meanwithstd{48.8}{1.1}\\

    \bottomrule
    \end{tabular}
    }
\end{table*}

\subsection{Multispectral and SAR Data} To verify that our contributions go beyond the RGB domain, we have also evaluated MaSoN on multispectral (MS) and SAR data. We use DEO~\cite{wolf2026deo} backbone for multispectral and CopernicusFM~\cite{wang2025copfm} for SAR. Most hyperparameters remain the same, except for backbone-dependent settings, for example, layer-ids (for swin we use all 4, and for ViT-based Cop-FM we use [3, 5, 7, 11]) and input normalisation parameters, which we take from the authors. The results for MS are shown in Table~\ref{atab:ms} and for SAR in the original paper. The model achieves superior performance compared to RGB models on OSCD-multispectral. The same holds for SAR, where only pixel-differencing and CVA directly support this modality. Other related methods cannot be easily extended to new modalities. This is especially restrictive in the SAM~\cite{kirillov2023sam} based methods, since SAM is a purely RGB foundation model.

\begin{table}[!h]
    \centering
        \caption{Performance with RGB/MS OSCD data}
    \label{atab:ms}
\setlength{\tabcolsep}{2pt}
    \begin{tabular}{lccc}
    \toprule
    	\multirow{2}{*}{~}& \multicolumn{3}{c}{OSCD} \\
& Pr. & Re. & F1\\
 \hline
Ours (RGB) & 34.2 & 60.3 & 43.5\\
Ours (MS) & 43.4 & 47.0 & 45.1\\

    \bottomrule
    \end{tabular}
\end{table}

\subsection{Different Noise Distribution} To verify how important the random distribution is from which we sample both the irrelevant and relevant noise, we have exchanged each with the Laplace random distribution. The results can be seen in Table~\ref{atab:dist}. The probability distribution plays an important role. Additionally, it is important to point out that all of these results are still above the previous state-of-the-art.

\begin{table*}[!h]
    \centering
        \caption{Different noise distributions ablation}
    \label{atab:dist}
    \resizebox{\linewidth}{!}{
\setlength{\tabcolsep}{2pt}
    \begin{tabular}{llccc|ccc|ccc|ccc|ccc|ccc}
    \toprule
    	\multirow{2}{*}{\makecell{Irrel. \\ noise}} & \multirow{2}{*}{\makecell{Rel. \\ noise}}& \multicolumn{3}{c}{SYSU} & \multicolumn{3}{c}{LEVIR} & \multicolumn{3}{c}{GVLM} & \multicolumn{3}{c}{CLCD} & \multicolumn{3}{c}{OSCD} & \multicolumn{3}{c}{\textit{Avg.}} \\
 && Pr. & Re. & F1& Pr. & Re. & F1& Pr. & Re. & F1& Pr. & Re. & F1& Pr. & Re. & F1& Pr. & Re. & F1\\
 \hline
Gauss & Gauss & \meanwithstd{72.3}{1.2} & \meanwithstd{52.3}{3.7} & \meanwithstd{60.6}{2.2} & \meanwithstd{25.9}{1.2} & \meanwithstd{78.5}{4.2} & \meanwithstd{38.9}{1.2} & \meanwithstd{43.2}{4.9} & \meanwithstd{78.5}{6.1} & \meanwithstd{55.4}{2.9} & \meanwithstd{46.3}{1.7} & \meanwithstd{66.4}{6.8} & \meanwithstd{54.3}{1.5} & \meanwithstd{34.2}{2.1} & \meanwithstd{60.3}{2.8} & \meanwithstd{43.5}{1.0} & \meanwithstd{44.4}{1.9} & \meanwithstd{67.2}{3.8} & \meanwithstd{50.6}{0.4}\\

Gauss & Laplace & \meanwithstd{52.2}{3.6} & \meanwithstd{67.2}{9.4} & \meanwithstd{58.3}{1.6} & \meanwithstd{16.0}{2.3} & \meanwithstd{87.9}{8.0} & \meanwithstd{26.9}{3.1} & \meanwithstd{15.5}{3.6} & \meanwithstd{88.7}{6.7} & \meanwithstd{26.1}{5.0} & \meanwithstd{21.7}{3.6} & \meanwithstd{82.2}{9.1} & \meanwithstd{34.0}{3.5} & \meanwithstd{28.9}{4.8} & \meanwithstd{69.0}{8.3} & \meanwithstd{40.2}{3.7} & \meanwithstd{26.9}{3.4} & \meanwithstd{79.0}{7.9} & \meanwithstd{37.1}{2.7}\\
Laplace & Gauss & \meanwithstd{64.2}{2.6} & \meanwithstd{66.8}{5.5} & \meanwithstd{65.3}{1.4} & \meanwithstd{20.1}{2.0} & \meanwithstd{93.0}{2.9} & \meanwithstd{32.9}{2.6} & \meanwithstd{17.4}{3.2} & \meanwithstd{94.9}{2.5} & \meanwithstd{29.2}{4.5} & \meanwithstd{30.1}{3.1} & \meanwithstd{85.5}{4.6} & \meanwithstd{44.3}{2.7} & \meanwithstd{24.8}{3.0} & \meanwithstd{76.9}{5.2} & \meanwithstd{37.3}{2.9} & \meanwithstd{31.3}{2.6} & \meanwithstd{83.4}{3.9} & \meanwithstd{41.8}{2.1}\\
Laplace & Laplace & \meanwithstd{76.5}{1.1} & \meanwithstd{28.8}{8.7} & \meanwithstd{41.3}{9.4} & \meanwithstd{29.3}{2.3} & \meanwithstd{48.9}{17.0} & \meanwithstd{35.6}{4.8} & \meanwithstd{47.6}{6.9} & \meanwithstd{58.3}{16.4} & \meanwithstd{50.6}{2.2} & \meanwithstd{49.7}{1.6} & \meanwithstd{45.3}{13.6} & \meanwithstd{46.3}{8.1} & \meanwithstd{47.0}{5.0} & \meanwithstd{41.8}{7.5} & \meanwithstd{43.6}{1.5} & \meanwithstd{50.0}{2.9} & \meanwithstd{44.6}{12.2} & \meanwithstd{43.5}{5.0}\\

    \bottomrule
    \end{tabular}
    }
\end{table*}

\section{Extended Implementation Details}
\label{a:ext_impl}

In this section, we present additional implementation details. Additional details for our model are in \Cref{asub:our_model}, for related unsupervised methods in \Cref{asub:related_unsup}, for supervised in \Cref{asub:related_sup}, and finally, ablation study experiments with additional details in \Cref{asub:ablation_impl}. For implementation details of computational efficiency experiments, refer to \Cref{a:comp} (including the discussion on computational overhead of noise generation strategy).

\subsection{Our model}
\label{asub:our_model}

We use pretrained weights from Huggingface in our experiments. The ViT-L pretrained with DINOv3~\cite{simeoni2025dinov3} is the following: \textcolor{weights}{facebook: dinov3-vitl16-pretrain-lvd1689m}. The main hyperparameters are listed in \Cref{par:impldet} of the main paper. Here we list some additional details. In the case of supervised learning, we train the model for 100 epochs instead of iterations, with a learning rate of $10^{-4}$ and a batch size of 32, while all other parameters stay the same. Rotation and flip augmentations are always applied with a chance of 30~\%. Each synthetic change is applied with a chance of 50~\%. The Perlin mask is sampled at full image dimension (256 x 256), thresholded at 0.5 and is then bilinearly downscaled to feature dimension (16x16 with DINOv3). All code is implemented with PyTorch and PyTorch Lightning. The metrics implementation comes from torchmetrics. All used code will be made available on GitHub upon acceptance.

\subsection{Unsupervised Related Methods}
\label{asub:related_unsup}

We use official code, provided by the authors, for all methods and just integrate our datasets. The following are versions of their code on GitHub:

\begin{itemize}
    \item DCVA~\cite{saha2019unsupervised}:  \textcolor{weights}{dcvaVHROptical, commit: e1d6af8365ca27062635cfa6fadee9198\-4302e1e}
    \item AnyChange~\cite{zheng2024anychange}:  \textcolor{weights}{pytorch-change-models, commit: 9c564feecfb577069b9a9bc\-5859264f768581046}
    \item Changen2~\cite{zgeng2025changen2}:  \textcolor{weights}{pytorch-change-models commit: 9c564feecfb577069b9a9bc\-5859264f768581046}
    \item CDRL~\cite{noh2022cdrl}:  \textcolor{weights}{CDRL commit: 9305057f00acfc1472673005c\-56949a90a2eb6b7}
    \item CDRL-SA~\cite{noh2024cdrlSA}:  \textcolor{weights}{CDRL-SA commit: a2e1ea203c455e28631bbca3bcbfc78f534\-20869}
    \item I3PE~\cite{chen2023_i3pe}:  \textcolor{weights}{I3PE commit: 3182c2918bd32a5b7dd44dc7ee71fe09ab92ed7b}
    \item HySCDG~\cite{benidir2025hyscdg}: \textcolor{weights}{HySCDG commit: d3e93feb7d4a59b723655186e47f9edb2de9\-acad}
    \item SCM~\cite{tan2024scm}: \textcolor{weights}{UCD-SCM commit: 72a6c3cf5c336604be4f6a8dc5e2\-401288e1779a}
    \item DynamicEarth~\cite{li2025dynearth}: \textcolor{weights}{DynamicEarth commit: c9ffd90cafbd791cd75a48a5717\-a902966c2436c}
    \item S2C~\cite{ding2025s2c}: \textcolor{weights}{S2C commit: 7bcec8658af25bfed87bc0be2109c7\-4e5106a677}
\end{itemize}

We keep all the hyperparameters the same as set by the authors. Since the weights for CycleGAN in CDRL are not available, we train our own with the official code using the default options.

\subsection{Supervised Related Methods}
\label{asub:related_sup}

\paragraph{Remote Sensing Foundation Models}

We use the official code and pre-trained weights provided by the authors for all remote sensing foundation models. The specific versions of the code used in our experiments can be accessed here:

\begin{itemize}
    \item CaCo~\cite{mall2023caco}: \textcolor{weights}{CACo commit: 7c4fbc8d9e65a4ef715596c5c275ab3ddb6a8193}
    \item SatMAE~\cite{cong2022satmae}: \textcolor{weights}{SatMAE commit: e31c11fa1bef6f9a9aa3eb49e8637c8b8952ba\-5e}
    \item GFM~\cite{mendieta2023gfm}: \textcolor{weights}{GFM commit: 4dd248e8544b3b6a49f5173b0931d97a17a7f424}
    \item MTP~\cite{wang2024mtp}: \textcolor{weights}{MTP commit: 962f7fd8781c095eb26db65ead3016e666b6d417}
\end{itemize}

Since foundation models don't have a predefined change detection architecture, we adopt the authors' architecture code and load the weights as the \textit{encoder} into our framework. The configuration is the same as for our model, with minor adjustments according to the authors' settings.

All methods use an initial learning rate of $10^{-4}$, except for the ViT-based models MTP and SatMAE, which follow the MTP~\cite{wang2024mtp} setup with a learning rate of $6\cdot10^{-5}$. Weight decay is set to $0.05$ for all methods, except for GFM, which uses $5 \cdot 10^{-4}$ as in~\cite{mendieta2023gfm}.
Feature extraction for CaCo and GFM is performed across all four hierarchical levels. For MTP and SatMAE, we follow~\cite{wang2024mtp} and extract features from layers with IDs 3, 5, 7, and 11 (MTP) and 7, 11, 15, and 23 (SatMAE). All methods use the UPerNet~\cite{xiao2018upernet} decoder, except ViT-based models, where UNet~\cite{ronneberger2015u} yields better performance and aligns with MTP’s original configuration. Other training settings, including Dice loss, cosine learning rate scheduling, and flip augmentations, remain consistent with our method.

\paragraph{Change Detection Specific Methods}

We use official code, provided by the authors, for all methods and just integrate our datasets. The following are versions of their code:

\begin{itemize}
    \item FCS-Diff~\cite{daudt2018fcn}:  \textcolor{weights}{fully\_convolutional\_change\_detection commit: 4dd83231f25\-319a7ebb16cbfa9912541ceabac9a}
    \item BIT~\cite{chen2021bit}:  \textcolor{weights}{BIT\_CD commit: adcd7aea6f234586ffffdd4e9959404f96271711}
    \item ChangeFormer~\cite{bandara2022changeFormer}:  \textcolor{weights}{ChangeFormer commit: afd1b7ed640aa265a2c730de9584\-16ae7356a2f9}
    \item BiFA~\cite{zhang2024bifa}:  \textcolor{weights}{BiFA commit: 56cd0da461e5e4b0d6a9b4f3321f0a81a91d21b8}
    \item SwinSUNet~\cite{zhang2022swinsunet}:  \textcolor{weights}{SwinSUNet commit: 721daf84238eda40fb49d626c21df4ed\-2246aa9e}
    \item ChangeMamba~\cite{chen2024changeMamba}: \textcolor{weights}{ChangeMamba commit: a91b82ee45059ce159f5f6f5d8e\-5818c33b84e68}
\end{itemize}

We keep all the hyperparameters the same as set by the authors, except the epoch number for BiFA and FC-Siam-Diff. For these two models, we halve the default number of epochs in the case of OSCD~\cite{daudt2018urban} dataset, since they start to overfit due to the dataset's small size.

\subsection{Ablation Study Implementation Details}
\label{asub:ablation_impl}

\paragraph{Sample Generation Strategy}

For the pixel-space irrelevant change generation experiment, we use the same images as the related method CDRL~\cite{noh2022cdrl} produced by learned transformation using CycleGan~\cite{zhu2017cycGan}. It is trained with official options and following code.

\paragraph{Backbone ablation}

The following architectures and weights from Huggingface are used for architecture ablation:
\begin{itemize}
    \item ViT-L DINOv3: \textcolor{weights}{facebook: dinov3-vitl16-pretrain-lvd1689m}
    \item ViT-L DINOv2: \textcolor{weights}{facebook: dinov2-large}
    \item ViT-B DINOv3: \textcolor{weights}{facebook: dinov3-vitb16-pretrain-lvd1689m}
    \item Swin: \textcolor{weights}{facebook: mask2former-swin-tiny-ade-semantic} 
    \item ResNet50: \textcolor{weights}{facebook: maskformer-resnet50-ade20k-full}
\end{itemize}

\paragraph{Multispectral and SAR ablation}

For multispectral encoder we use DEO~\cite{wolf2026deo} Swin B. For SAR, we use CopernicusFM~\cite{wang2025copfm} \textcolor{weights}{CopernicusFM\_ViT\_base\_varlang\_\-e100.pth} from \textcolor{weights}{wangyi111: Copernicus-FM}. We use the authors' normalisation in both cases. The rest of the architecture remains the same.

\paragraph{Extension to change filtering}

We integrate MaSoN as the change comparer in the pipeline of DynamicEarth~\cite{li2025dynearth} as a simple drop in replacement of raw DINOv2/v3. This means that the first (mask proposal) and last (filtering) phase remain unchanged, but we keep only the proposed masks that have at least 30\% of overlap with masks predicted by MaSoN. We also use the same filtering prompts: "building" for buildings and "background" for everything else.